\documentclass[preprint,12pt,authoryear]{elsarticle}
\usepackage[font=normalfont,labelfont=bf]{caption}

\usepackage{graphicx}
\usepackage{float}
\usepackage{siunitx}
\usepackage{stackengine}

\usepackage[usenames, dvipsnames]{xcolor}
\usepackage{colortbl}

\definecolor{darkcyan}{rgb}{0.0, 0.55, 0.55}
\usepackage[colorlinks=true]{hyperref}

\usepackage{tikz}
\usetikzlibrary{positioning}
\usetikzlibrary{arrows}
\usetikzlibrary{tikzmark}
\usetikzlibrary{backgrounds}
\usetikzlibrary{arrows.meta}
\usetikzlibrary{shadows} \usetikzlibrary{fadings}
\usetikzlibrary{shapes.arrows}
\usetikzlibrary{patterns}

\usepackage{amsmath,amsfonts,amssymb}
\usepackage{dsfont}
\usepackage{bm}

\usepackage{algorithm}
\usepackage{algpseudocode}

\usepackage{array}
\usepackage{diagbox}
\usepackage{multirow}
\usepackage{makecell}
\usepackage[export]{adjustbox}
\usepackage{subcaption}
\usepackage{cleveref}
\usepackage{arydshln}

\usepackage{pifont}
\newcommand{\cmark}{\ding{51}}%
\newcommand{\xmark}{\ding{55}}%
\DeclareMathOperator*{\argmin}{arg\,min}

\DeclareRobustCommand{\augiefamily}{%
\fontfamily{augie}\fontseries{m}\fontshape{n}\selectfont}
\DeclareTextFontCommand{\textaugie}{\augiefamily}
\DeclareMathAlphabet{\mathsfit}{T1}{\sfdefault}{\mddefault}{\sldefault}

\journal{Engineering Applications of AI}

\begin{document}

\begin{frontmatter}


\title{Unrecognizable Yet Identifiable: Image Distortion with Preserved Embeddings}

\author[karazin_math]{Dmytro Zakharov}
\ead{zamdmytro@gmail.com}

\author[macerata,eCampus]{Oleksandr Kuznetsov\corref{cor}}
\ead{kuznetsov@karazin.ua}
\cortext[cor]{Corresponding author}

\author[macerata]{Emanuele Frontoni}
\ead{emanuele.frontoni@unimc.it}

\affiliation[karazin_math]{
            organization={Department of Applied Mathematics, V.N. Karazin Kharkiv National University},
            addressline={4 Svobody Sq.},
            city={Kharkiv},
            postcode={61022},
            country={Ukraine}}
            
       
\affiliation[macerata]{
            organization={Department of Political Sciences, Communication and International Relations, University of Macerata},
            addressline={Via Crescimbeni, 30/32},
            city={Macerata},
            postcode={62100},
            country={Italy}}

\affiliation[eCampus]{
            organization={Faculty of Engineering, eCampus University},
            addressline={Via Isimbardi 10},
            city={Novedrate (CO)},
            postcode={22060},
            country={Italy}}



\begin{abstract}

Biometric authentication systems play a crucial role in modern security systems. However, maintaining the balance of privacy and integrity of stored biometrics derivative data while achieving high recognition accuracy is often challenging. Addressing this issue, we introduce an innovative image transformation technique that effectively renders facial images unrecognizable to the eye while maintaining their identifiability by neural network models, which allows the distorted photo version to be stored for further verification. While initially intended for biometrics systems, the proposed methodology can be used in various artificial intelligence applications to distort the visual data and keep the derived features close. By experimenting with widely used datasets \textit{LFW} and \textit{MNIST}, we show that it is possible to build the distortion that changes the image content by more than 70\% while maintaining the same recognition accuracy. We compare our method with previously state-of-the-art approaches. We publically release the source code\footnote{\textcolor{purple}{\url{https://github.com/ZamDimon/distortion-generator/tree/v1.0.0}}}.

\end{abstract}




\end{frontmatter}

\section{Introduction}\label{section:introduction}

In the digital age, one cannot overstate the necessity for robust cybersecurity systems. With the rapid integration of digital identities and the increasing reliance on virtual platforms for many activities, safeguarding personal and organizational data has become a crucial problem \citep{ai_for_biometric}. This surge in digitalization has simultaneously amplified cybersecurity vulnerabilities, making exploring innovative and effective security solutions essential. One such solution is biometric-based authentication systems \citep{authentication_techniques}, which can remove the need to memorize passwords and provide an additional security layer in authentication systems.

Current biometric systems, while revolutionary in many respects, have their drawbacks. A huge concern is the risk of irreversible compromise; once a biometric trait is exposed or stolen, it is compromised forever, unlike traditional passwords or tokens that can be easily changed \citep{vulnerabilities_biometric}. Furthermore, issues like data privacy, susceptibility to spoofing attacks, and the challenge of maintaining high accuracy under varied conditions underscore the limitations of existing biometric technologies. Integrating these systems into diverse platforms also presents challenges in terms of scalability, interoperability, and user accessibility.

Against this backdrop, many research studies have explored how biometric data can be effectively managed to prevent revealing a person's identity. Obviously, storing the original biometrics data, like fingerprint or face image, is entirely insecure; once the attacker gets access to the database, he knows the biometrics of each person registered in the system. Therefore, there should be a way to store the derived features, which can help identify the person without revealing as much information as possible.

In this paper, we propose a novel Non-Distortive Cancelable Biometrics system that addresses these challenges by leveraging advanced machine learning techniques to derive secure, revocable, and privacy-preserving biometric templates. Our approach differs from traditional cancelable biometrics methods in that it allows direct comparison between unaltered probe samples and transformed reference templates, thereby enhancing both the security and the usability of the authentication process. Through rigorous experimental evaluations on the benchmark \textit{LFW} \citep{lfw} facial dataset and the \textit{MNIST} \citep{mnist} handwritten digit dataset, we demonstrate the effectiveness of our system in terms of recognition accuracy, template security, and revocability. We also provide detailed ablation studies to analyze the impact of various design choices and hyperparameters on the system's performance. Furthermore, we situate our work within the broader context of biometric security research and highlight its unique contributions and advantages over existing state-of-the-art methods.

\subsection{Our Contribution}
The \textit{primary objective} of this research is to explore and validate the feasibility of an image distortion technique while preserving the features. Our approach diverges from traditional methods by avoiding the distortion of original biometric data, instead employing advanced Artificial Intelligence (AI) algorithms for data analysis and template generation.

\begin{figure*}
\begin{center}
\begin{tabular}{ccccccc}
& \textbf{Real} & \textbf{Generated} & \textbf{Real} & \textbf{Generated} & \textbf{Real} & \textbf{Generated} \\ 
 & \includegraphics[width=.1\linewidth,valign=m]{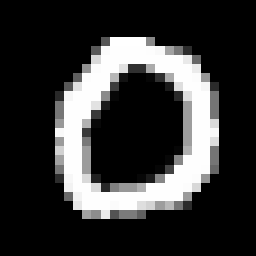} & \includegraphics[width=.1\linewidth,valign=m]{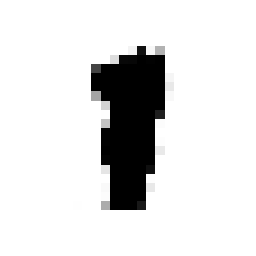} & \includegraphics[width=.1\linewidth,valign=m]{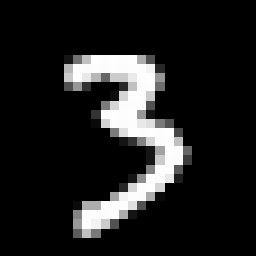}& \includegraphics[width=.1\linewidth,valign=m]{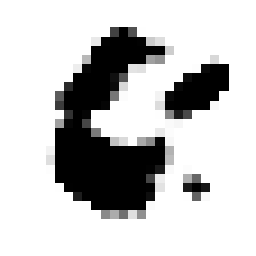} & \includegraphics[width=.1\linewidth,valign=m]{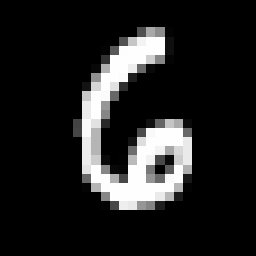} & \includegraphics[width=.1\linewidth,valign=m]{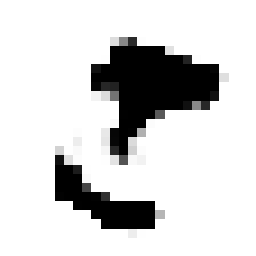} \\
\textsf{MNIST} & \includegraphics[width=.1\linewidth,valign=m]{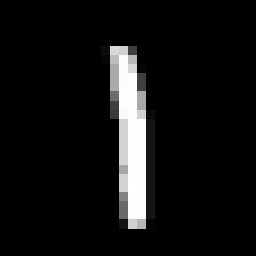} & \includegraphics[width=.1\linewidth,valign=m]{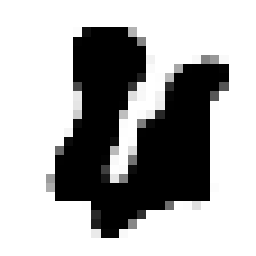} & \includegraphics[width=.1\linewidth,valign=m]{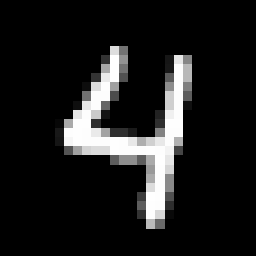}& \includegraphics[width=.1\linewidth,valign=m]{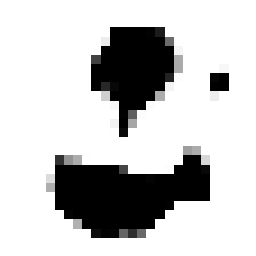} & \includegraphics[width=.1\linewidth,valign=m]{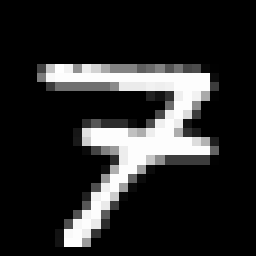} & \includegraphics[width=.1\linewidth,valign=m]{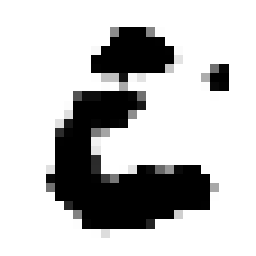} \\
& \includegraphics[width=.1\linewidth,valign=m]{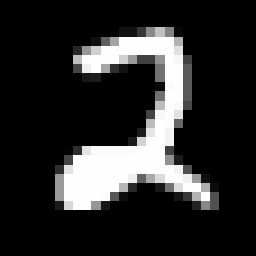} & \includegraphics[width=.1\linewidth,valign=m]{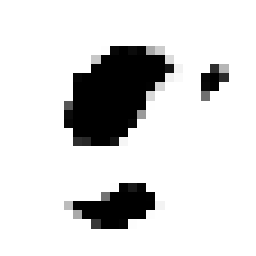} & \includegraphics[width=.1\linewidth,valign=m]{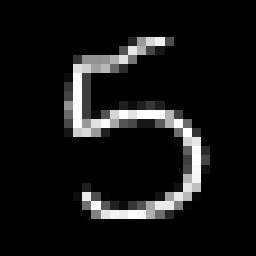}& \includegraphics[width=.1\linewidth,valign=m]{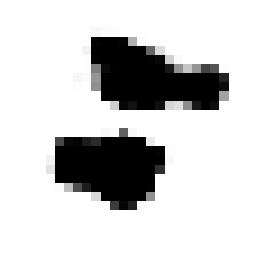} & \includegraphics[width=.1\linewidth,valign=m]{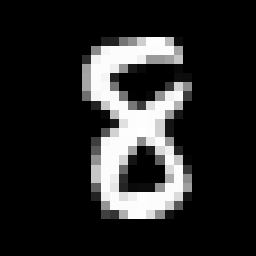} & \includegraphics[width=.1\linewidth,valign=m]{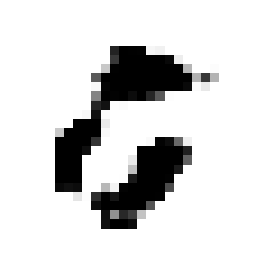}\\
\\ 
 & \includegraphics[width=.1\linewidth,valign=m]{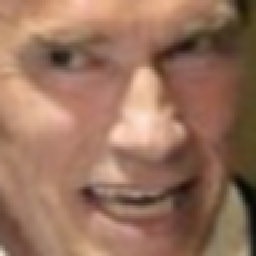} & \includegraphics[width=.1\linewidth,valign=m]{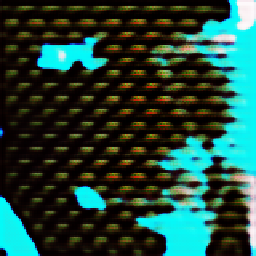} & \includegraphics[width=.1\linewidth,valign=m]{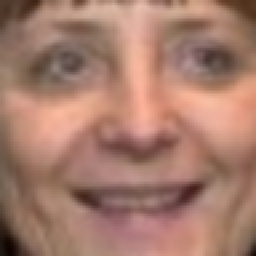}& \includegraphics[width=.1\linewidth,valign=m]{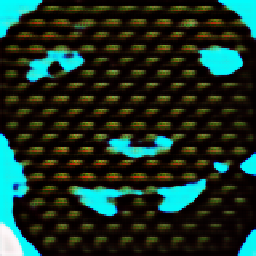} & \includegraphics[width=.1\linewidth,valign=m]{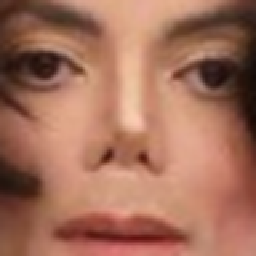} & \includegraphics[width=.1\linewidth,valign=m]{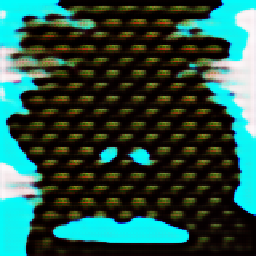} \\
\textsf{LFW} & \includegraphics[width=.1\linewidth,valign=m]{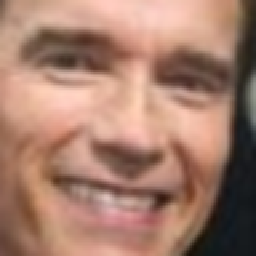} & \includegraphics[width=.1\linewidth,valign=m]{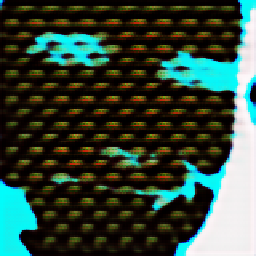} & \includegraphics[width=.1\linewidth,valign=m]{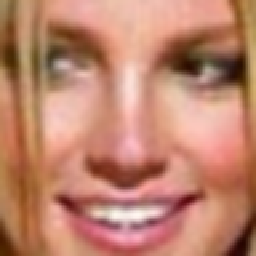}& \includegraphics[width=.1\linewidth,valign=m]{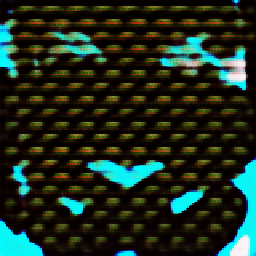} & \includegraphics[width=.1\linewidth,valign=m]{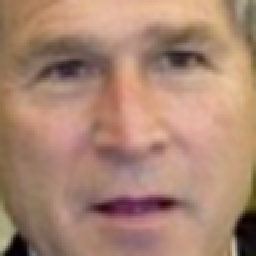} & \includegraphics[width=.1\linewidth,valign=m]{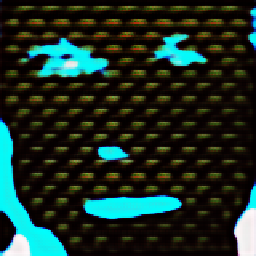} \\
& \includegraphics[width=.1\linewidth,valign=m]{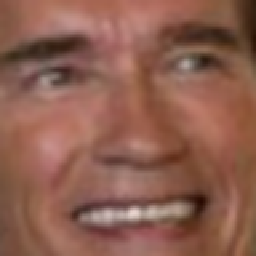} & \includegraphics[width=.1\linewidth,valign=m]{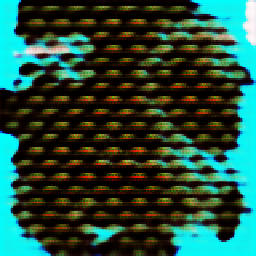} & \includegraphics[width=.1\linewidth,valign=m]{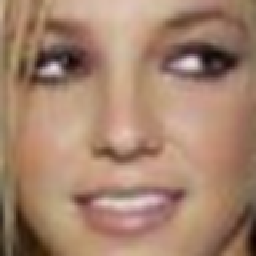}& \includegraphics[width=.1\linewidth,valign=m]{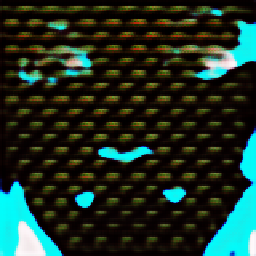} & \includegraphics[width=.1\linewidth,valign=m]{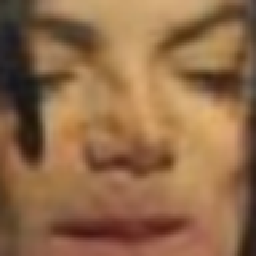} & \includegraphics[width=.1\linewidth,valign=m]{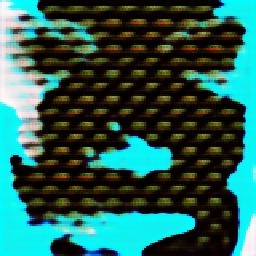}
\end{tabular}
\end{center}
\caption{An example of using our proposed image distortion technique on images from \textit{MNIST} \citep{mnist} and \textit{LFW} \citep{lfw} datasets. While authentic and generated images significantly differ, the feature vectors of both images in pairs are relatively close.}
\label{table:examples}
\end{figure*}

For better clarity, consider the pairs depicted in \autoref{table:examples}. While the images on the left can be easily recognized and identified, the photos on the right almost do not reveal any information about the underlying image. However, in contrast to, say, the cryptographic hashing function, the pictures on the right can be compared with the initial image via specified comparison algorithm that the distortion generator referenced while training. 

Our methodology involves an analysis of biometric data integrity and the application of state-of-the-art AI techniques. We utilize the idea of \textit{Triplet Networks} to develop a sophisticated metric for biometric data comparison, ensuring the security of the data while maintaining its original characteristics. The research encompasses a series of experiments using the \textit{MNIST} \citep{mnist} and \textit{LFW} \citep{lfw} dataset to validate our system's effectiveness empirically.

This research contributes significantly to the field of biometric security. By introducing a non-distortive approach to cancelable biometrics, we provide a solution that balances the need for safety with the imperative of protecting individual privacy. Our findings could influence future developments in biometric authentication, paving the way for more secure and privacy-conscious systems. Moreover, since we essentially conceal the original image, the distorted version can be publicly revealed and potentially used in cryptographic protocols\footnote{Note that for guaranteeing high security, a much more extensive cryptoanalysis is needed, which we leave for further studies.}.

The paper is structured as follows: after this introduction, we delve into the theoretical framework in \autoref{section:Background}, experimental methodology in \autoref{section:methods}, implementation details in \autoref{section:implementation}, results analysis in \autoref{section:results}, and a comprehensive discussion of our research's implications and future directions in \autoref{section:comparison} and \autoref{section:conclusion}.

\subsection{State-of-the-Art}\label{section:State of the Art}

The field of biometric security has witnessed significant advancements in recent years, with numerous studies exploring various approaches to ensure the privacy, security, and operational efficiency of biometric systems. This section provides an overview of the most relevant and influential works that have shaped the landscape of cancelable biometrics and motivated our research on Non-Distortive Cancelable Biometrics.

We begin by examining the foundational work of \citet{Bansal_Garg_2022}, who introduced a cancelable biometric template protection scheme combining format-preserving encryption with Bloom filters. Their approach laid the groundwork for enhancing security while maintaining recognition performance. However, their primary focus on the encryption aspect left room for further exploration of the operational challenges involved in deploying such systems across diverse platforms. Our research aims to bridge this gap by leveraging AI algorithms to simplify operational complexities.

Building upon this foundation, \citet{Helmy_El-Shafai_El-Rabaie_El-Dokany_El-Samie_2022} proposed a novel hybrid encryption framework based on Rubik's cube technique for cancelable biometric systems. Their innovative method for securing multi-biometric systems showcased the potential for advanced encryption techniques in this domain. Nevertheless, their emphasis on encryption raised questions about the ease of integration and scalability. Our study addresses these concerns by proposing a more holistic approach that achieves security without compromising system architecture simplicity.

Moving beyond encryption, \citet{Kauba_Piciucco_Maiorana_Gomez-Barrero_Prommegger_Campisi_Uhl_2022} delved into the practical aspects of cancelable biometrics for finger vein recognition. Their analysis of three different approaches and their impact on recognition performance and security provided valuable insights into the challenges and opportunities in this specific biometric modality. However, their focus on finger vein recognition highlighted the need for a more comprehensive framework applicable to a wider range of biometric types. Our research responds to this need by proposing a versatile AI-driven metric suitable for various biometric modalities.

\citet{Nayar_Thomas_Emmanuel_2021} introduced a graph-based approach for secure cancelable palm vein biometrics, offering a novel perspective on template security. While their method demonstrated the potential of graph-based techniques, its specificity to palm vein biometrics limited its direct applicability to other domains. Recognizing the importance of a more universal solution, our approach is designed to be adaptable across different biometric systems.

\citet{Yang_Wang_Kang_Johnstone_Bedari_2022} made significant strides in cancelable fingerprint authentication with their linear convolution-based system. Their work underscored the critical importance of safeguarding fingerprint template data. However, their focus on fingerprints left an opportunity for exploration in other biometric modalities. Our research seizes this opportunity by proposing a comprehensive solution that can be readily adapted to various biometric types.

The innovative application of partial Hadamard transform to cancelable biometrics by \citet{Wang_Deng_Hu_2017} marked a significant milestone in the field. Their method enhanced the security of binary biometric representations and effectively prevented the reconstruction of original data. Building upon their groundbreaking work, our research integrates AI algorithms, expands the scope to multiple biometric modalities, and emphasizes the preservation of original data integrity, thereby addressing a crucial gap in user-friendly and secure biometric authentication.

\citet{Yang_Wang_Shahzad_Zhou_2021} tackled the vulnerability of traditional random projection-based cancelable biometrics to attack via record multiplicity (ARM). While their feature-adaptive random projection method enhanced security against this specific type of attack, there remained a need for a more comprehensive approach encompassing broader security concerns. Our research fills this void by introducing a holistic framework that ensures high recognition accuracy while addressing a wider range of security risks.

The biometrics-based secure key agreement protocols proposed by \citet{Akdogan_Karaoglan_Altop_Eskandarian_Levi_2018} showcased the importance of integrating cancelability into biometric data. Their work, particularly the SKA-CB protocol, highlighted the potential of cancelable biometrics in enhancing security. However, their focus on key agreement protocols left room for exploration in terms of operational flexibility and cross-platform adaptability. Our research addresses these broader aspects, offering a more versatile solution applicable to diverse biometric applications.

\citet{Kaur_Khanna_2020} made valuable contributions to privacy and security in network/cloud-based remote biometric authentication by combining cancelable pseudo-biometric identities with secret sharing. While their approach tackled key security concerns, its emphasis on remote authentication indicated an opportunity for improvement in local system integration and broader biometric modalities. Our research bridges these gaps by proposing a system that delivers effective performance in both local and remote contexts, across a wide spectrum of biometric types.

The iris-based cancelable biometric cryptosystem introduced by \citet{Kausar_2021} showcased the potential of combining biometrics with symmetric key cryptography for securing healthcare data on smart cards. While their work provided valuable insights into biometric data security in healthcare, the focus on iris biometrics and healthcare applications underscored the need for a more generalized approach applicable across different sectors. Our research addresses this need by offering a generalizable and adaptable solution in the form of Non-Distortive Cancelable Biometrics.

\citet{Lee_Teoh_Uhl_Liang_Jin_2021} proposed a tokenless cancellable biometrics scheme for multimodal biometric systems, emphasizing biometric template protection without relying on tokens. While their approach innovated in enhancing security and simplifying the authentication process, it did not fully address the operational complexities related to system integration across various platforms. Our study aims to provide a comprehensive solution that simplifies integration and operational aspects in diverse application scenarios.

\citet{Murakami_Ohki_Kaga_Fujio_Takahashi_2019}  made significant contributions to fast and secure biometric identification with their cancelable biometric scheme based on correlation-invariant random filtering. While their approach showcased innovation in security and computational efficiency, its primary target was large-scale identification systems. Our research complements their work by offering a scalable and adaptable solution catering to both large-scale and individualized biometric authentication needs.

\citet{Yang_Wang_Hu_Zheng_Valli_2018} explored the potential of cancelable multi-biometric systems by combining fingerprint and finger-vein biometrics. Their approach underscored the importance of feature-level fusion for enhanced recognition accuracy and security. However, their focus on fingerprint and finger-vein biometrics highlighted an opportunity for a more expansive framework. Our research builds upon their work by developing a framework applicable to a wider range of biometric modalities, enhancing versatility and applicability in diverse scenarios.

In summary, the cited studies represent a carefully curated selection of the most influential and relevant works in the field of cancelable biometrics. Each study has made significant contributions to the advancement of biometric security, addressing specific challenges and proposing innovative solutions. However, their individual focus on specific aspects, modalities, or applications has left room for a more comprehensive, scalable, and adaptable approach. Our research on Non-Distortive Cancelable Biometrics aims to bridge these gaps, drawing inspiration from the strengths of these studies while addressing their limitations. By providing a holistic solution that enhances security, operational efficiency, and applicability across various biometric modalities and application scenarios, our work represents a significant step forward in the field of biometric security.

\section{Background}\label{section:Background}

Since this section introduces the math for formal algorithms and techniques specification, we enumerate the notation used in \autoref{table:notation}. Any other notation will be introduced in the course of explaining the methodology.

\begin{table}
    \centering
    \caption{Notation used in the paper}
    \begin{tabular}{cc:cc}
        \Xhline{3.5\arrayrulewidth}
        $x,\alpha$ & scalar variables & $\mathbf{x},\boldsymbol{X}$ & vectors and matrices \\
        $n$ & dataset size & $\mathbb{P}$ & probability\\
        $m$ & feature vector size & $\mathbb{E}$ & expected value \\
        $n_B$ & batch size & $\mathsfit{B}$ & batch of images \\
        $X$ & image & $\mathsfit{T}$ & a set of image triplets \\
        $d$ & distance & $\theta$ & trainable parameters \\
        $\ell$ & pointwise loss & $\mathcal{I}$ & set of images \\
        $\|\cdot\|_1,\|\cdot\|_2$ & $L_1$ and $L_2$ norm & $\mathcal{F},\mathcal{G}$ & neural network functions \\
        $f(x \mid \mu)$ & function of $x$ parameterized by $\mu$ & $f \circ g(x)$ & composite function $f(g(x))$
        \\
        $\odot$ & pointwise multiplication & $\star$ & convolution operation
        \\
        \Xhline{3.5\arrayrulewidth}
    \end{tabular}
    \label{table:notation}
\end{table}

\subsection{Image Distortion Techniques}
\begin{figure*}
\centering
  \begin{subfigure}{.5\textwidth}
    \includegraphics[width=.9\linewidth]{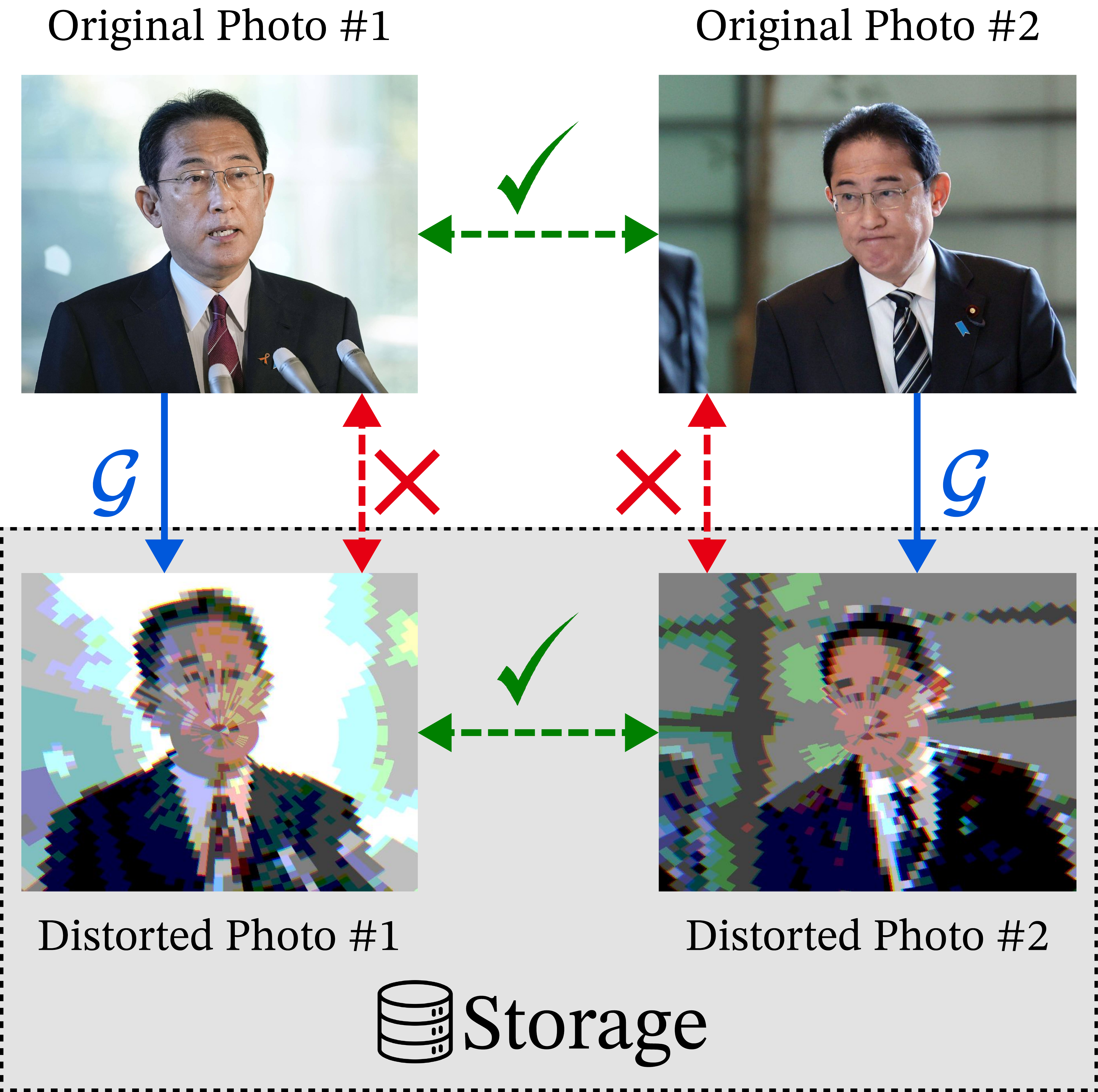}
    \caption{Cancelable Biometrics}
  \end{subfigure}%
  \begin{subfigure}{.5\textwidth}
    \includegraphics[width=.9\linewidth]{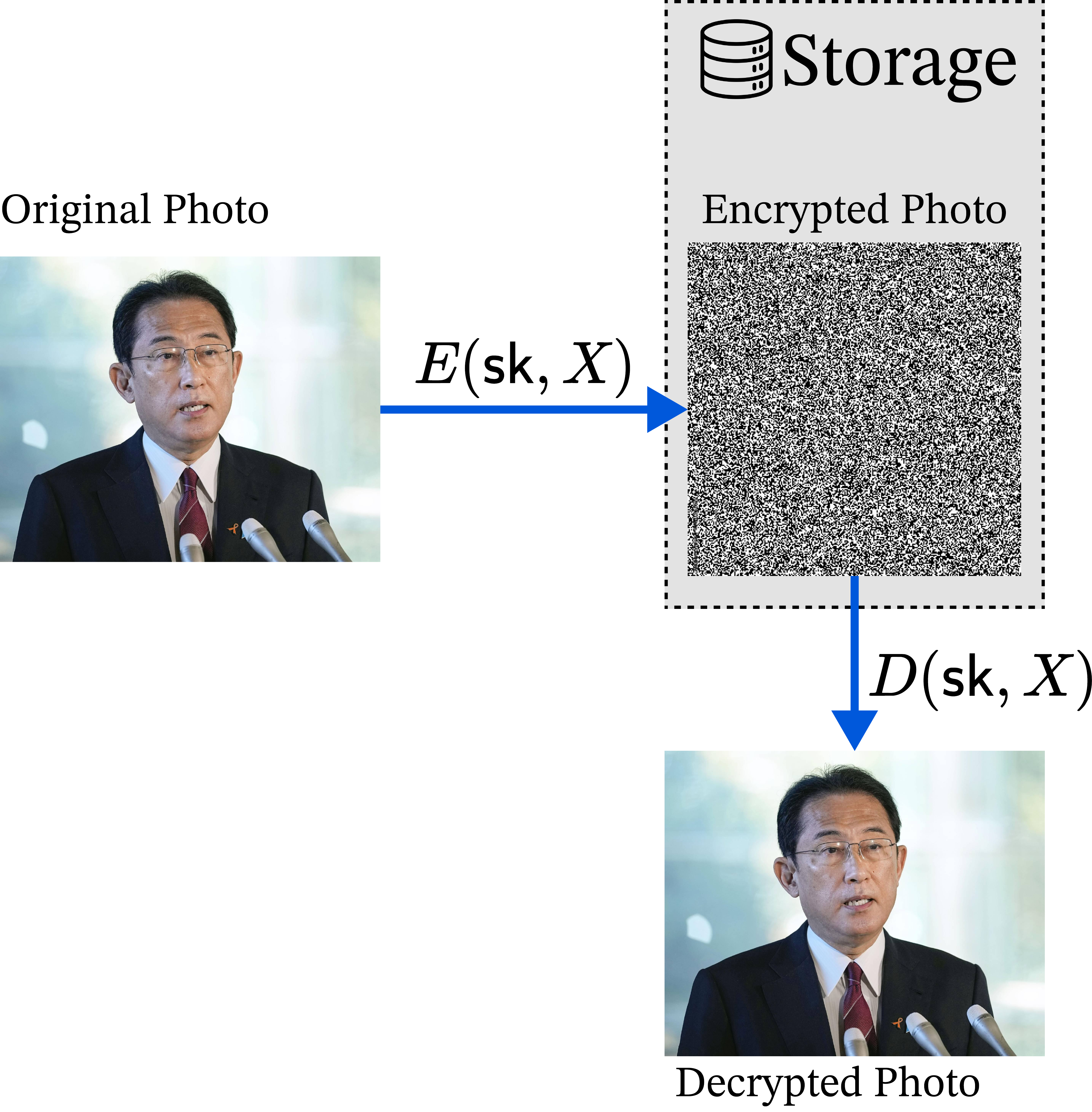}
    \caption{Image Encryption}
  \end{subfigure}
\caption{Two primary methods of storing biometric data: (a) -- storing distorted templates, (b) -- encryption an image via the secret key $\mathsf{sk}$.}
\label{fig:visualization}
\end{figure*}

Although our primary focus was on cancelable biometrics in the previous section, it is worth specifying other approaches of biometrics security used in the literature and seeing what our proposed solution brings to the table. Currently, based on the comprehensive survey by \citet{encoding_image_3}, image security research resolves around the following primary topics, visualized in \autoref{fig:visualization}:
\begin{enumerate}
    \item \textbf{Steganography:} Hiding information inside the cover image that is unrecognizable for a human eye \citep{steganography_overview,hiding_image}. Currently, as seen in \citep{steganography_overview}, many neural network architectures can resolve this problem.
    \item \textbf{Cancelable Biometrics:} The uninvertible image conversion $\mathcal{G}$ into the unrecognizable representation that can be further compared with another similarly converted image directly: see case (a) in \autoref{fig:visualization}. 
    \item \textbf{Image encryption:} The process of deterministic image transformation to the unrecognizable representation (cipher) via the secret key $\mathsf{sk}$ that is further decodable to a trusted party by the same key $\mathsf{sk}$. In this case, two encrypted images cannot be compared without decrypting them \citep{encoding_image_3,encoding_image_1,encoding_image_2}. The process is illustrated in case (b), \autoref{fig:visualization}.
\end{enumerate}

While the first topic is exciting, we aim not to hide the information inside another image for two primary reasons: (1) typically, not much data can be handled by steganography, and (2) the security against an active attacker is lower compared to the latter two methods. That being said, our study is best related to the second and third topics, so we focus on a comparison of these two subjects. We discuss the advantages of our approach by considering an example of building the most simplistic authorization system, which all these methods are intended for in the first place.

\begin{figure}
    \centering
    \includegraphics[width=0.7\textwidth]{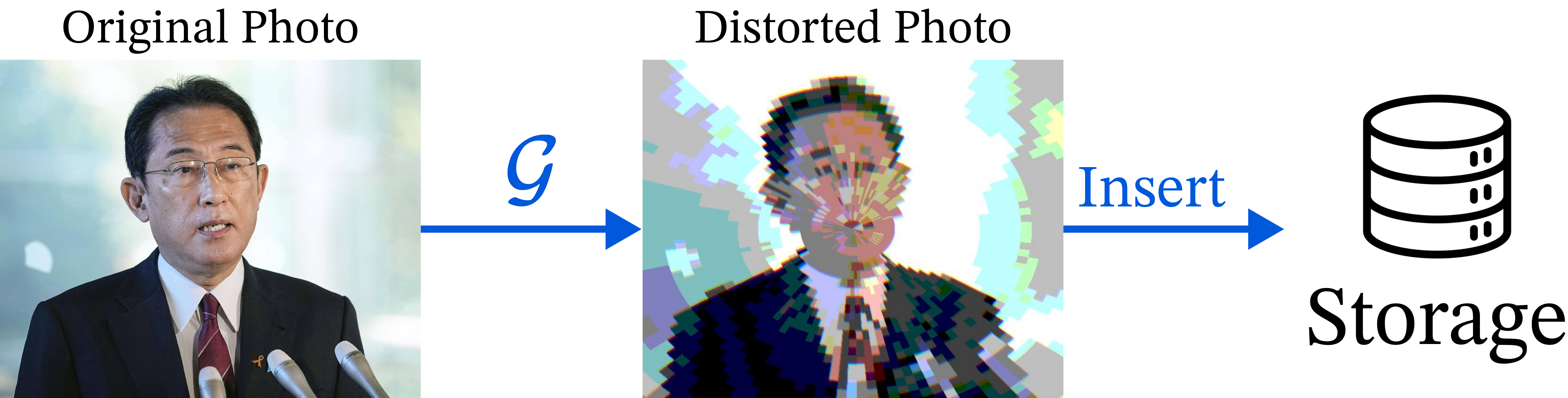}
    \caption{Facial recognition feature registration flow.}
    \label{fig:registration}
\end{figure}

First, consider the flow of a \textit{user registration}, depicted in \autoref{fig:registration}. Suppose some user wants to use a facial recognition authorization feature on account. In that case, the system must be provided with the biometrics data $X$, which is then processed by some generator function. 

\textit{Cancelable biometrics} converts this image to another representation via the function $\mathcal{G}$ that will look almost the same if the same person takes another photo. Formally, if $X_1$ and $X_2$ are the photos of the same person, $\mathcal{G}$ must be hard to invert and $\mathcal{G}(X_1) \approx \mathcal{G}(X_2)$.

\textit{Image encryption} will convert an image to a cryptographically secure cipher $c_X=E(\mathsf{sk},X)$, which can be further decoded using the same secret key via decryption function $D(\mathsf{sk},c_X) = X$. 

Either way, the processed data from $X$ gets saved in the database, which we call a ``template'' data $T$. 

Now, consider the setting where a user passes an image $X$, and the system wants to verify that the user exists in the database. Specifically, we need an algorithm to output $1$ if $X$ belongs to the same person as some template $T$ from the database and $0$ to different people. Here comes the main difference between the methods considered.

\begin{figure*}
\centering
  \begin{subfigure}{.5\textwidth}
    \includegraphics[width=.9\linewidth]{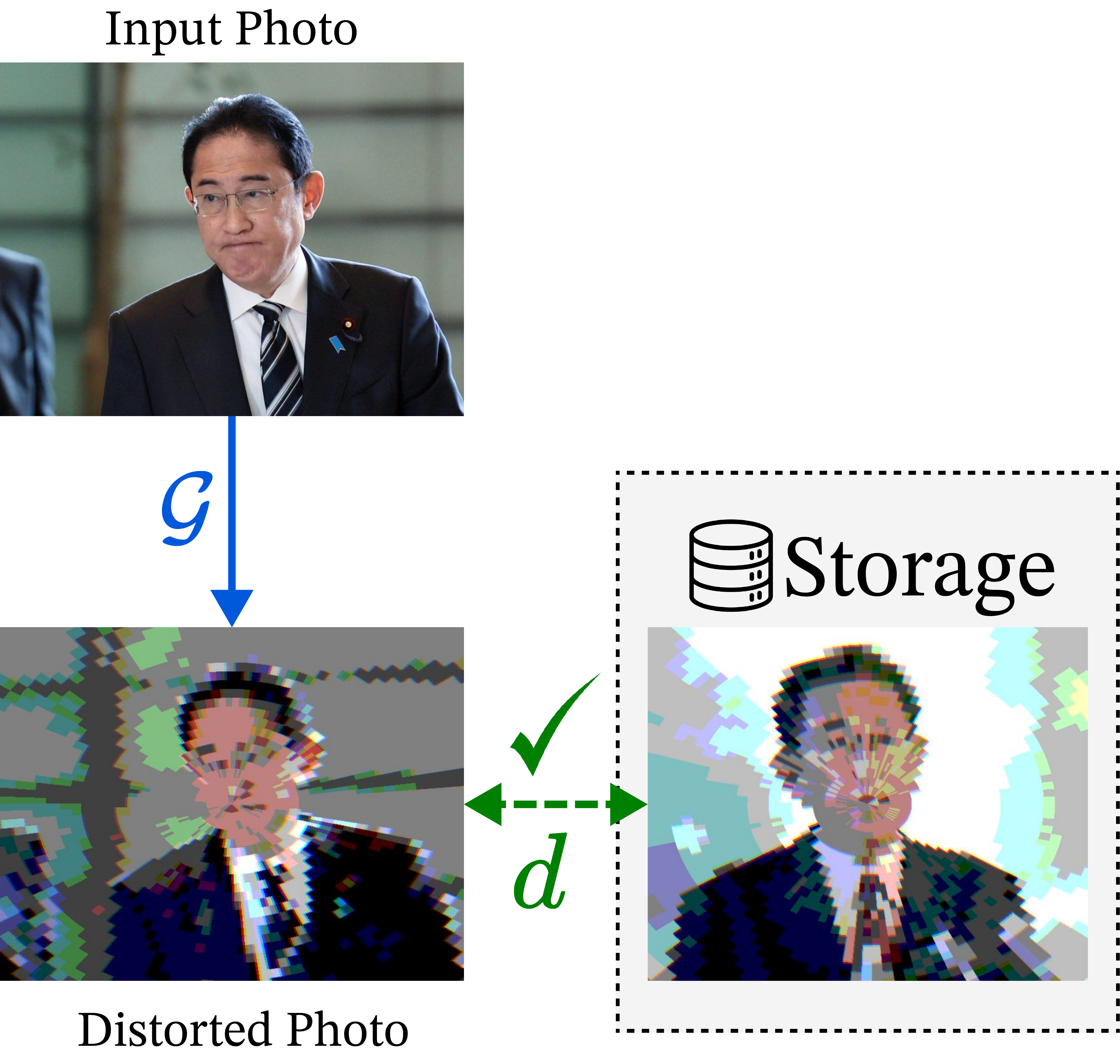}
    \caption{Cancelable Biometrics}
  \end{subfigure}%
  \begin{subfigure}{.5\textwidth}
    \includegraphics[width=.9\linewidth]{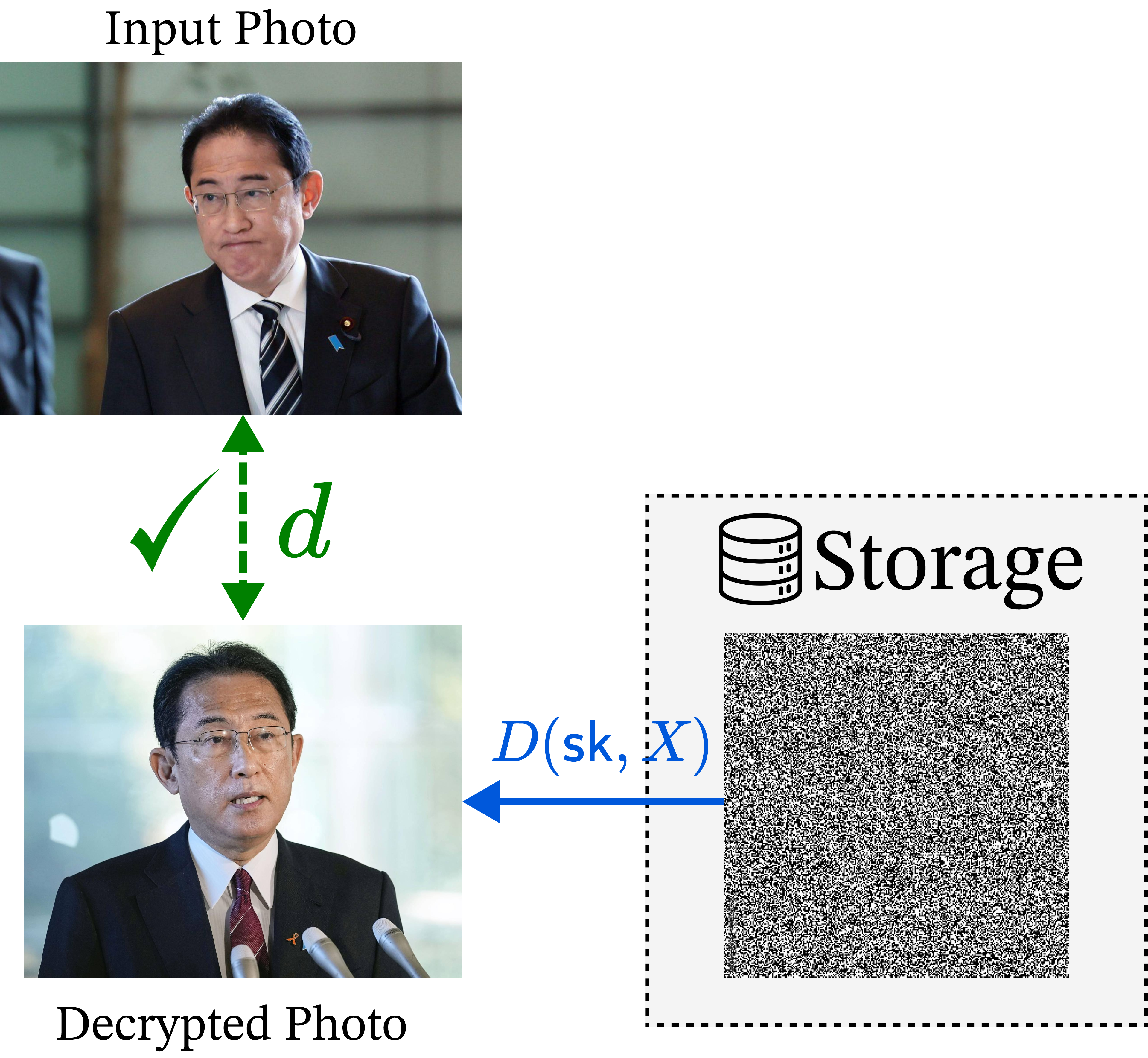}
    \caption{Image Encoding}
  \end{subfigure} \\ \vspace{20px}
  \begin{subfigure}{.5\textwidth}
    \includegraphics[width=.9\linewidth]{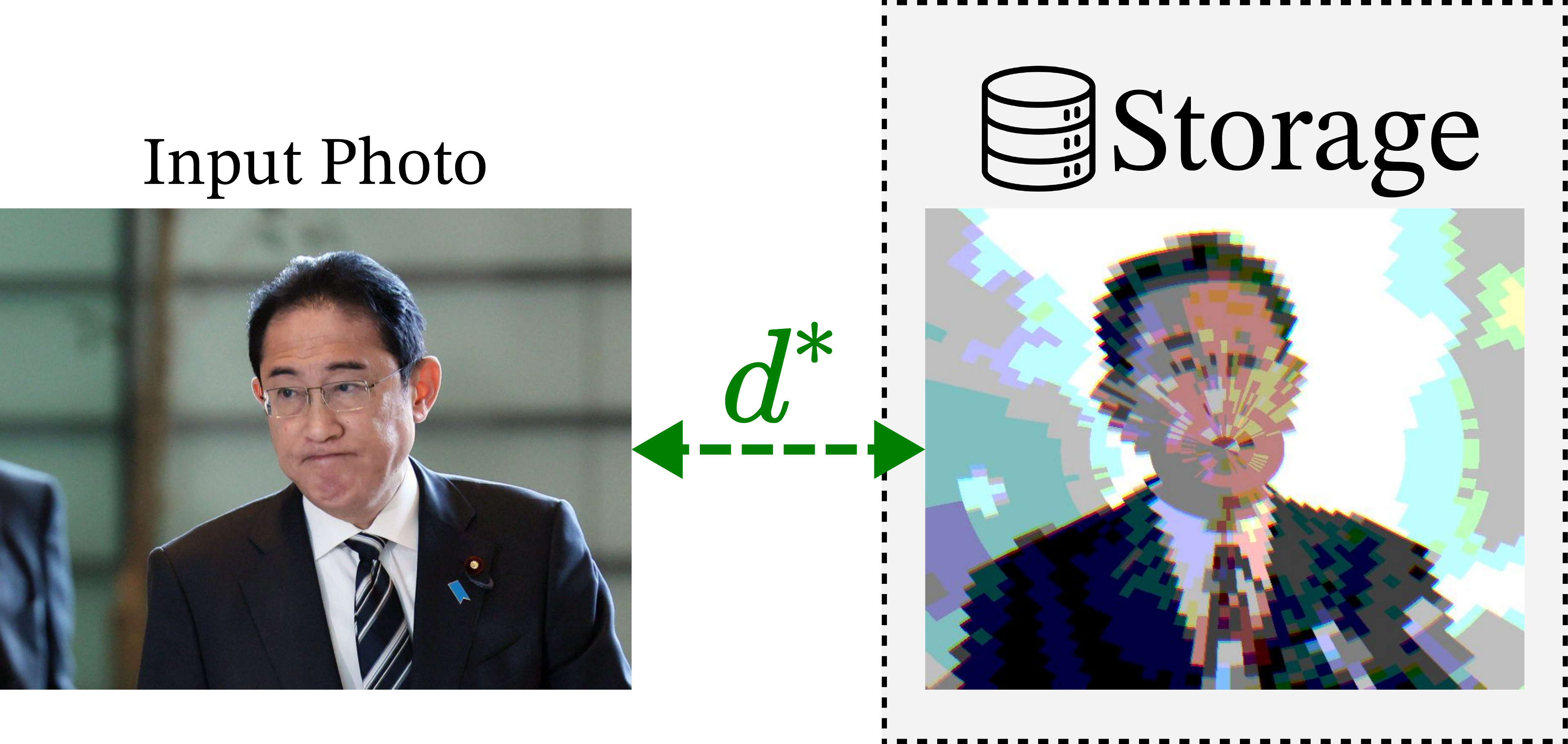}
    \caption{Our login user registration flow}
  \end{subfigure}%
\caption{Comparison of different login flows using biometrics data, where $\mathcal{G}$ denotes the generation function, $D(\mathsf{sk},\star)$ is a decryption function with a secret key $\mathsf{sk}$, $d$ denotes the traditional image distance metrics, while $d^*$ -- a secret one. In flow \textit{(a)}, we first generate and then compare a distorted image with a template. In flow \textit{(b)}, we find the inverse of a template and compare it with an input. In our proposed flow \textit{(c)}, we compare template and image directly.}
\label{fig:login}
\end{figure*}

Consider cases \textit{(a)} and \textit{(b)} in \autoref{fig:login}, where we depict the login flows for cancelable biometrics and image encryption-based approaches, respectively. In both cases, we use a metric of image comparison, or in our particular case, image distance $d$. As the simplest example, $d$ might be an Euclidean or Hamming metric or can encapsulate some complex mechanism such as an image-matching technique. Now, we explain the flows for each of the cases:

\begin{itemize}
    \item In \textit{cancelable biometrics} approach we need to firstly generate an image $\mathcal{G}(X)$ and compare it with $T$ using distance $d$. If $d(\mathcal{G}(X), T)$ is small enough, we consider ownership of both $X$ and $T$ to be the same.
    \item In \textit{image encryption} approach, we retrieve the original image from the template $D(\mathsf{sk},T)$ and compare it to $X$. If $d(D(\mathsf{sk},T), X)$ is small enough, we again consider the ownership of $X$ and $T$ to be the same.
\end{itemize}

\subsection{Advantages of Our Solution}
The proposed Non-Distortive Cancelable Biometrics system offers several key advantages over traditional biometric authentication techniques, particularly in terms of security, privacy, and operational efficiency. Table \ref{table:comparison} provides a comprehensive comparison of our approach with cancelable biometrics and biometric encryption methods across multiple evaluation criteria.

One of the primary benefits of our system lies in its ability to perform single-step comparison between the unaltered probe image $X$ and the distorted reference template $T$. This is achieved through the use of a secret comparison metric $d^*$ that is unknown to external entities. By directly computing $d^*(X,T)$ without the need for any intermediate image transformations, our approach significantly reduces the computational overhead and latency associated with the authentication process. In contrast, both cancelable biometrics and biometric encryption techniques typically require a two-step procedure, involving the application of a transformation function $\mathcal{G}(X)$ or a decryption operation $D(\mathsf{sk},T)$, followed by a comparison using a standard metric $d$. The elimination of these additional steps not only streamlines the authentication workflow but also enhances the overall system efficiency and scalability.

Moreover, our Non-Distortive Cancelable Biometrics system effectively conceals the original biometric information, without compromising the recognition accuracy. By leveraging advanced machine learning techniques to derive a highly discriminative yet visually dissimilar representation of the biometric data, our approach ensures that the stored reference templates $T$ reveal minimal information about the original input $X$. This property is formally quantified by the large difference between $X$ and $T$ under traditional comparison metrics $d$, such as Euclidean distance or Hamming distance. Consequently, even if an attacker gains access to the stored templates, it would be computationally infeasible to reconstruct the original biometric data, thereby providing a strong assurance of user privacy and mitigating the risk of permanent biometric compromise.

Another salient advantage of our system is its ability to maintain or even surpass the recognition accuracy of non-protected biometric comparison, as will be demonstrated in Section \ref{section:results}. This is in stark contrast to conventional cancelable biometrics methods, which often incur a noticeable degradation in recognition performance due to the application of irreversible transformations to the biometric data. Prior studies, such as those by \citet{Maiorana2010}, \citet{Rathgeb2014OnAO}, and \citet{Takahashi2009}, have consistently reported a decrease in accuracy when employing cancelable biometrics techniques. By preserving the discriminative power of the original biometric representations while ensuring their security and privacy, our approach strikes an optimal balance between the competing objectives of recognition performance and template protection.

Furthermore, our Non-Distortive Cancelable Biometrics system eliminates the need for secure key management, which is a critical requirement in biometric encryption schemes. In such methods, the secret key $\mathsf{sk}$ plays a pivotal role in the encryption and decryption of the biometric templates. Consequently, the security of the entire system hinges on the proper management and protection of these keys. Any compromise or unauthorized access to the secret keys would render the encrypted templates vulnerable, allowing an attacker to recover the original biometric data. In contrast, our approach achieves template protection without relying on any secret information, thereby obviating the need for complex key management infrastructures and reducing the attack surface of the biometric system.

The non-invertibility and revocability of the protected templates are two essential requirements for any biometric template protection scheme. Our Non-Distortive Cancelable Biometrics system satisfies both these criteria, as evidenced by the irreversible nature of the learned mapping between the input biometric image $X$ and its secure template $T$. The non-invertibility property ensures that, given a protected template, it is computationally infeasible to recover the original biometric data. This is achieved by the use of one-way transformation functions that are designed to be resistant to inversion attacks. Moreover, our approach supports the revocability and renewability of templates, allowing for the generation of multiple independent protected templates from the same biometric input. In the event of a template compromise, the affected template can be easily revoked and replaced with a new one, without the need for re-enrolling the user or changing the underlying biometric data. This flexibility is essential for maintaining the long-term security and reliability of the biometric system.

\begin{table}[htbp]
\centering
\caption{Comparison of biometric template protection approaches}
\label{table:comparison}
\begin{tabular}{l|c|c|c}
\Xhline{3.5\arrayrulewidth}
\textbf{Criteria} & \textbf{Cancelable} & \textbf{Biometric} & \textbf{Non-Distortive} \\
& \textbf{Biometrics} & \textbf{Encryption} & \textbf{Cancelable} \\
\Xhline{3.5\arrayrulewidth}
Non-Invertibility & \textcolor{ForestGreen}{\cmark} & \textcolor{ForestGreen}{\cmark} & \textcolor{ForestGreen}{\cmark}\\
Revocability & \textcolor{ForestGreen}{\cmark} & \textcolor{ForestGreen}{\cmark} & \textcolor{ForestGreen}{\cmark}\\
Accuracy Preservation & \textcolor{Red}{\xmark} & \textcolor{ForestGreen}{\cmark} & \textcolor{ForestGreen}{\cmark}\textcolor{ForestGreen}{\cmark}\\
Key Management & \textcolor{ForestGreen}{\cmark} & \textcolor{Red}{\xmark}\textcolor{Red}{\xmark} & \textcolor{ForestGreen}{\cmark}\\
Matching Complexity & \textcolor{Red}{\xmark} & \textcolor{Red}{\xmark} & \textcolor{ForestGreen}{\cmark}\textcolor{ForestGreen}{\cmark}\\
Compatibility & \textcolor{Red}{\xmark} & \textcolor{ForestGreen}{\cmark} & \textcolor{ForestGreen}{\cmark}\\
\Xhline{3.5\arrayrulewidth}
\end{tabular}
\end{table}

Table \ref{table:comparison} presents a comprehensive comparison of our Non-Distortive Cancelable Biometrics approach with traditional cancelable biometrics and biometric encryption techniques. The evaluation criteria encompass key aspects such as non-invertibility, revocability, accuracy preservation, key management complexity, matching complexity, and compatibility with existing biometric systems.
As evident from the table, our approach excels in all the considered criteria, demonstrating its superiority over the other template protection methods. The non-invertibility and revocability properties are successfully achieved, ensuring the security and renewability of the protected templates. Moreover, our system maintains or even enhances the recognition accuracy compared to non-protected biometric matching, as denoted by the double tick (\textcolor{ForestGreen}{\cmark}\textcolor{ForestGreen}{\cmark}) in the corresponding row. This is a significant advantage over cancelable biometrics methods, which often suffer from accuracy degradation due to the application of irreversible transformations.

Another notable strength of our approach lies in its simplified key management and reduced matching complexity. By eliminating the need for secret keys and enabling direct comparison between the probe and reference templates, our system minimizes the operational overhead and enhances the efficiency of the authentication process. This is in contrast to biometric encryption techniques, which require careful key management and involve computationally intensive encryption and decryption operations.

Furthermore, our Non-Distortive Cancelable Biometrics system is designed to be compatible with existing biometric recognition frameworks and infrastructures. This compatibility facilitates the seamless integration of our template protection method into practical biometric systems, without necessitating significant modifications or adaptations. Such interoperability is crucial for the wide-scale adoption and deployment of our approach in real-world applications.
In summary, the proposed Non-Distortive Cancelable Biometrics system offers a comprehensive and effective solution for biometric template protection, addressing the limitations of previous approaches. By achieving non-invertibility, revocability, and accuracy preservation, while simplifying key management and matching complexity, our method paves the way for secure, efficient, and privacy-preserving biometric authentication. The comparative analysis presented in Table \ref{table:comparison} highlights the superior performance and practical advantages of our approach, positioning it as a promising candidate for the next generation of biometric security systems.

\subsection{Triplet Loss Usage for training an Embedding Model} \label{section:embedding}

Denote by $\mathcal{I}$ a set of images. To further avoid confusion with terminology, we define the term \textit{embedding model} as the function $\mathcal{F}: \mathcal{I} \to \mathbb{R}^m$, which maps an image to a low-dimensional representation in $\mathbb{R}^m$, sometimes called a \textit{feature vector}.

The embedding model is an excellent tool for various problems, not only in terms of computational efficiency but also since we can encapsulate core patterns in data using only hundreds of numbers (instead of ten thousands of them). For example, consider papers \citep{geoposition_1} and \citep{geoposition_2}, where embeddings store information about geographical position (for more examples, see \autoref{section:triplet_network}).

Similarly to \textit{FaceNet} paper by \citet{facenet}, we limit the output to the unit hypersphere $S^{m-1} = \{\mathbf{x} \in \mathbb{R}^m: \|\mathbf{x}\|_2 = 1\}$ with the embedding size of $m$. This step is optional, though: in fact, any function $\mathcal{F}$ might be provided, not limited to deep learning ones, as long as the gradient descent algorithm can be applied.

The main purpose of our neural network is to create ``similar'' embeddings for images from the same class and ``different'' for ones from different classes. We define the measure of ``distinctiveness'' as follows:
\begin{equation}\label{eq:emb_distance}
d_{\mathcal{F}}(X,Y) = \|\mathcal{F}(X)-\mathcal{F}(Y)\|_2^2.
\end{equation}
This way, if $X_1$ and $X_2$ belong to the same class, while $Y$ to a different one, $d_{\mathcal{F}}(X_1,X_2)$ must be much smaller than both $d_{\mathcal{F}}(X_1,Y)$ and $d_{\mathcal{F}}(X_2,Y)$. 

However, the neural network must know how to learn to produce such embeddings. For that reason, we consider the dataset $\mathsfit{T} = \{(A_i,P_i,N_i)\}_{i=0}^{n}$ of size $n$, where $A_i$ and $P_i$ are images from the same class (called \textit{anchor} and \textit{positive} images, respectively) whereas $N_i$ from a different one (called \textit{negative} image). 

The idea of triplet loss is to constrain an embedding of an anchor image $A$ to be closer to the corresponding embedding of $P$ than an image $N$ by a positive value $\mu$ (called \textit{margin}). So ideally, for all triplets $(A,P,N) \in \mathsfit{T}$ we want:
\begin{equation}
d_{\mathcal{F}}(A,P) < d_{\mathcal{F}}(A,N) - \mu
\end{equation}
Surely, this is practically hard to achieve. Therefore, from the probabilistic perspective, suppose that we take random samples $(A,P,N)$ from a true distribution $p_{\text{data}}$. Our goal is to maximize the probability of the aforementioned relationship by picking the following extractor $\hat{\mathcal{F}}$:
\begin{equation}
    \hat{\mathcal{F}} = \max_{\mathcal{F}}\left\{\mathbb{P}_{(A,P,N) \sim p_{\text{data}}}\left[d_{\mathcal{F}}(A,P) < d_{\mathcal{F}}(A,N) - \mu\right]\right\}
\end{equation}
Again, since solving this problem directly is complicated (although one can find the detailed probabilistic analysis in \citet{warburg2021bayesian}), the following loss function is considered, which is called a \textit{triplet loss function}:
\begin{equation}\label{eq:triplet_loss}
\ell(A,P,N \mid \mathcal{F})=\mathsf{ReLU}\left(d_{\mathcal{F}}(A,P) - d_{\mathcal{F}}(A,N) + \mu\right),
\end{equation}
where the $\mathsf{ReLU}(x) = \max\{0, x\}$ is defined as usual. Then, we minimize the expected error $\mathbb{E}_{(A,P,N) \sim p_{\text{data}}}[\ell(A,P,N \mid \mathcal{F})]$ to get the estimate of optimal $\mathcal{F}$. 

\subsection{Triplet Network}\label{section:triplet_network}
Triplet loss and triplet neural networks play a crucial role in many areas of computer vision: for instance, they are used in face recognition \citep{facenet,normface}, person reidentification \citep{person_reidentification}, object tracking \citep{object_tracking}, and even generative neural networks \citep{triplet_generative}. 

\begin{figure}
    \centering
    \includegraphics[width=\textwidth]{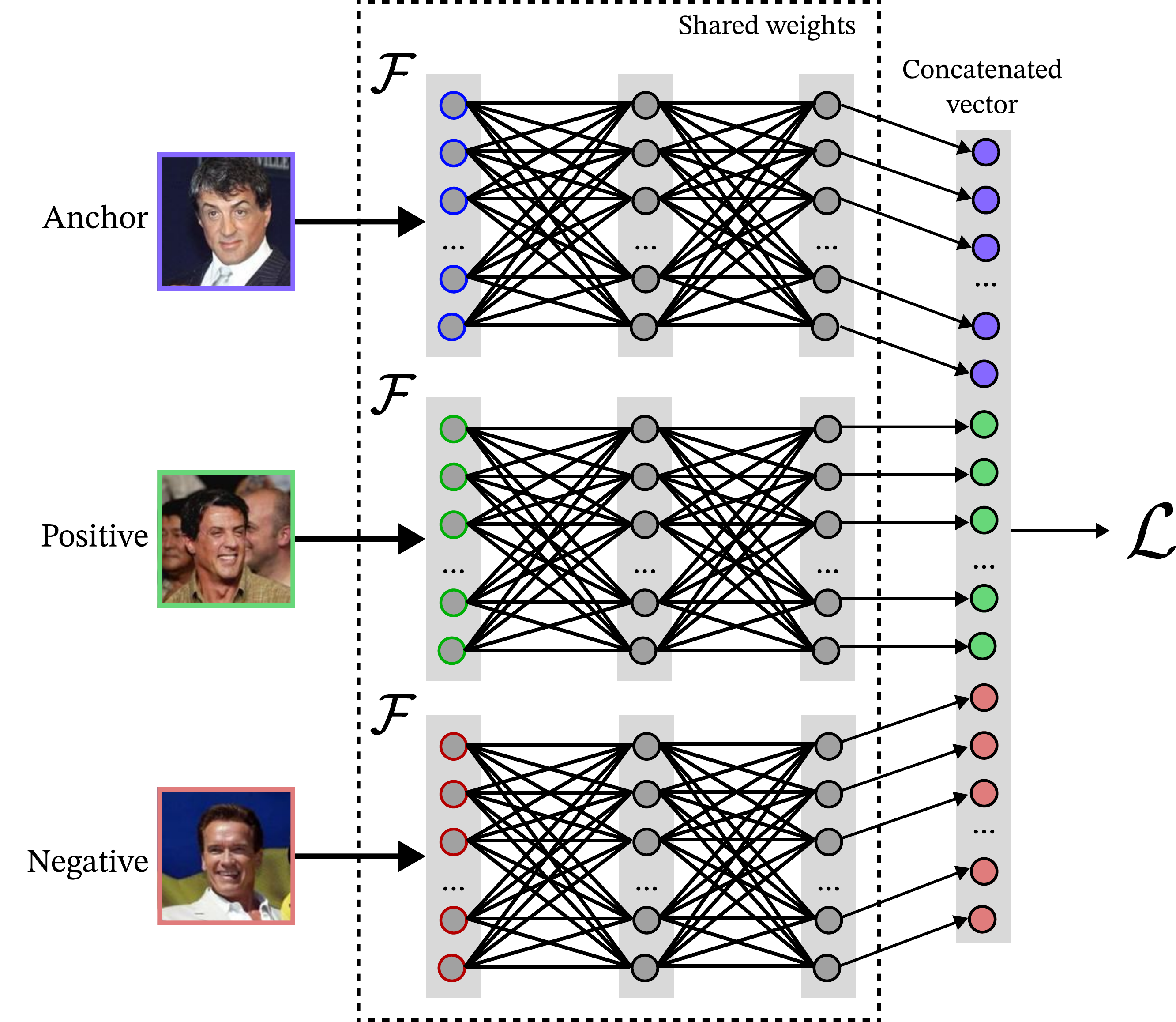}
    \caption{Triplet Network architecture. We input three images (anchor, positive, and negatives), then, using embedding model $\mathcal{F}$ with shared parameters, retrieve three feature vectors and concatenate them to get the loss value.}
    \label{fig:triplet_network}
\end{figure}

To examine the structure of a triplet neural network, we refer to \autoref{fig:triplet_network}.

Triplet Network uses three copies of an embedding model with shared parameters \citep{triplet_network}. Using the triplet loss defined in \autoref{section:embedding}, we calculate the loss and update the weights of an embedding model. We can then safely retrieve and use the embedding model for our purposes. Specifically, the most basic example algorithm is outlined in \autoref{alg:training_triplet}. 

\begin{algorithm}
\caption{The simplest training algorithm of embedding model using triplet network architecture.}
\label{alg:training_triplet}
\begin{algorithmic}
\State \textbf{\textsf{Input:}} Triplet dataset $\mathsfit{T} = \{(A_i,P_i,N_i)\}_{i=1}^{n}$ of size $n$, batch size $n_B \ll n$, learning rate $\eta$, and the initial neural network parameterization $\theta^{\langle 0 \rangle}$.
\State \textbf{\textsf{Output:}} Learned parameterization $\theta$, minimizing the expected loss on $\mathsfit{T}$.
\State \textbf{\textsf{Training:}}
\For{each batch $\mathsfit{B} := \{(A_i,P_i,N_i)\}_{i=1}^{n_B} \subset \mathsfit{T}$}
    \State 1. Find embedding vectors $
    \boldsymbol{a}_i \gets \mathcal{F}(A_i), \boldsymbol{p}_i \gets \mathcal{F}(P_i), \boldsymbol{n}_i \gets \mathcal{F}(N_i)$ for each $i = 1,\dots,n_B$
    \State 2. Find the batch loss $\mathcal{L}(\theta) \gets \sum_{i=1}^{n_B} \ell(\boldsymbol{a}_i,\boldsymbol{p}_i,\boldsymbol{n}_i \mid \theta)\big/n_B$ where 
    \[
    \ell(\boldsymbol{a}_i,\boldsymbol{p}_i,\boldsymbol{n}_i \mid \theta) = \|\boldsymbol{a}_i-\boldsymbol{p}_i\|_2^2-\|\boldsymbol{a}_i-\boldsymbol{n}_i\|_2^2+\mu.
    \]
    \State 3. Update $\theta$ using the gradient descent. In its simplest form, we use 
    \[
    \theta^{\langle j+1\rangle} \gets \theta^{\langle j\rangle} - \eta \nabla_{\theta}\mathcal{L}(\theta^{\langle j\rangle}).
    \]
\EndFor
\end{algorithmic}
\end{algorithm}

\section{Methods}\label{section:methods}

\subsection{Overview}
Distortion generator is a function $\mathcal{G}: \mathcal{I} \to \mathcal{I}$, which generates a distorted image from a given one. This generator must meet the following two criteria:
\begin{enumerate}
    \item Difference between images $\mathcal{G}(X)$ and $X$ is as large as possible. We call the metrics for such difference $d_{\text{img}}: \mathcal{I} \times \mathcal{I} \to \mathbb{R}_{\geq 0}$.
    \item Difference between embeddings $\mathcal{F}\circ \mathcal{G}(X)$ and $\mathcal{F}(X)$ is as small as possible. We call the metrics for this difference $d_{\text{emb}}: \mathbb{R}^m \times \mathbb{R}^m \to \mathbb{R}_{\geq 0}$.
\end{enumerate}

\begin{figure}
    \centering
    \includegraphics[width=0.4\textwidth]{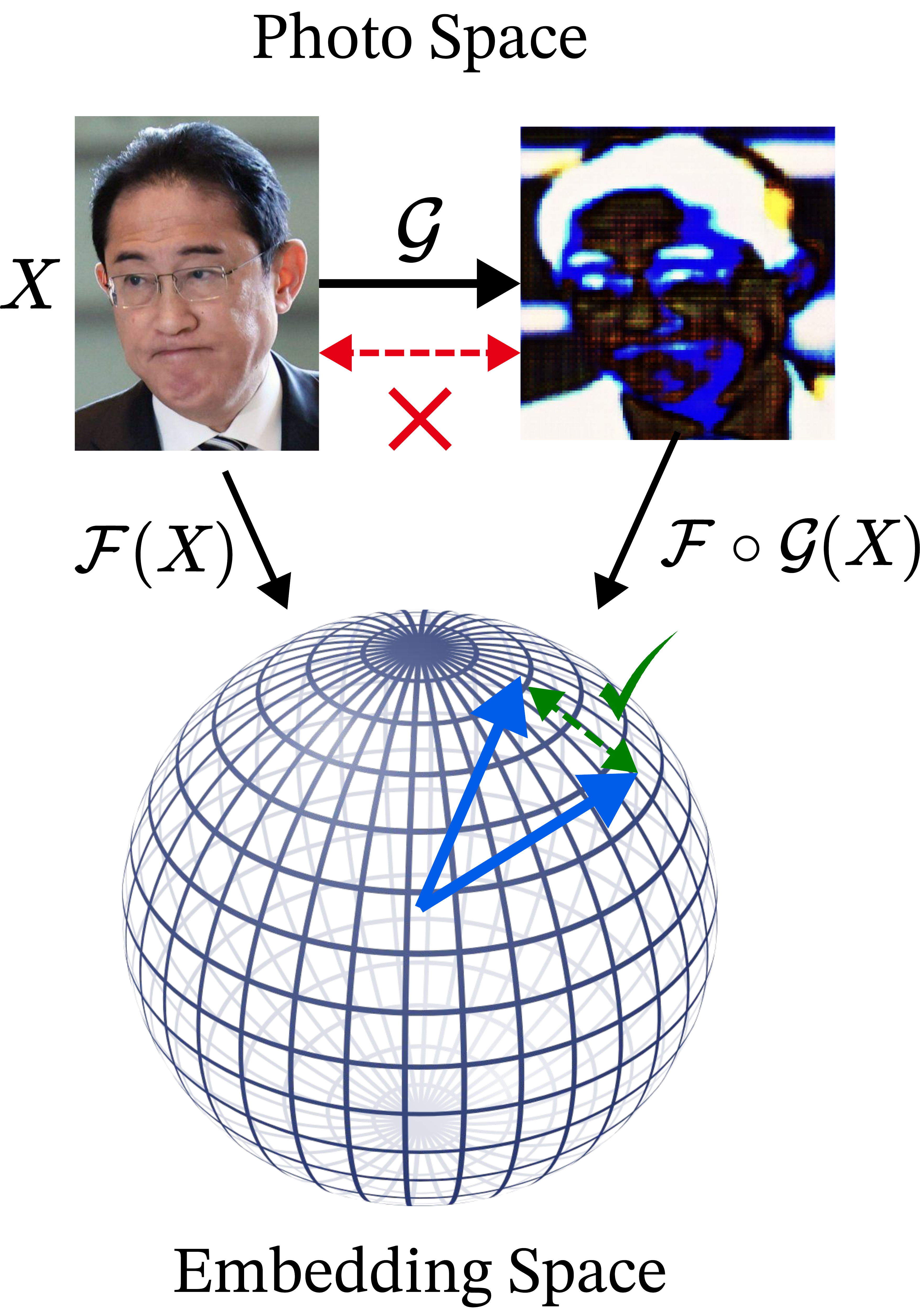}
    \caption{Illustration of an optimization problem: while $d_{\text{img}}$ must be large in the photo space, the distance between embeddings $d_{\text{emb}}$ must be small.}
    \label{fig:space_visualization}
\end{figure}

These two conditions are informally illustrated in \autoref{fig:space_visualization}.

Suppose inputs are taken from the true distribution $p_{\text{data}}$. This way, informally, we want to have:
\begin{gather}\label{eq:informal_optimization}
\max_{\mathcal{G}} \mathbb{E}_{X \sim p_{\text{data}}}\left[d_{\text{img}}(\mathcal{G}(X), X)\right] \\ \text{while} \; \; \min_{\mathcal{G}} \mathbb{E}_{X \sim p_{\text{data}}}\left[ d_{\text{emb}}(\mathcal{F} \circ \mathcal{G}(X), \mathcal{F}(X))\right]
\end{gather}

Note that in this case, we cannot employ the idea of a two-player minimax game used in GAN \citep{gan} directly since we cannot modify the embedding neural network $\mathcal{F}$, although this idea does seem attractive at first glance. 

However, if we wanted to train a pair $(\mathcal{F},\mathcal{G})$ together, that could be possible. That is an excellent topic for future research, but for now, we restrict ourselves $\mathcal{F}$ to be fixed.

\subsection{Loss Function}\label{section:loss}

To represent the optimization problem above, we define the following loss function for a single image:
\begin{equation}\label{eq:loss_trainer}
\ell(X \mid \mathcal{G}, \mathcal{F}) = (1-\pi_{\text{emb}}) \cdot \ell_{\text{img}}(X\mid\mathcal{G}) + \pi_{\text{emb}}\cdot\ell_{\text{emb}}(X\mid\mathcal{G},\mathcal{F}),
\end{equation}
where $\pi_{\text{emb}} \in [0,1]$ is a positive hyperparameter, regulating the importance of $\ell_{\text{emb}}$ in contrast to $\ell_{\text{img}}$. 

We define the two loss components as follows:
\begin{align}
&\ell_{\text{img}}(X \mid \mathcal{G}) = -d_{\text{img}}(\mathcal{G}(X), X),\\ &\ell_{\text{emb}}(X\mid \mathcal{G},\mathcal{F}) = \mathsf{ReLU}(d_{\text{emb}}(\mathcal{F}\circ \mathcal{G}(X), \mathcal{F}(X))-\alpha).
\end{align}

Note that $\ell_{\text{img}}$ is always negative since we want to \textit{maximize} the difference between images. Also, we decide to use $\mathsf{ReLU}(d_{\text{emb}}(\cdot)-\alpha)$ for $\ell_{\text{emb}}$ instead of $d_{\text{emb}}(\cdot)$ since otherwise neural network might focus primarily on reducing the distance between embeddings. However, if we use the $\mathsf{ReLU}$ function, we do not punish the neural network for an embedding difference unless it exceeds $\alpha$. In this sense, $\alpha$ also serves as a parameter that regulates how well we want our generator to fit embeddings: the larger $\alpha$ is, the more distinct images are according to metrics $d_{\text{img}}$, but less similar according to $d_{\text{emb}}$ (see \autoref{section:margin_tuning}).

Let us now choose the concrete expressions for distances. We use $d_{\mathcal{F}}$ from \autoref{eq:emb_distance} for the embedding difference:
\begin{equation}
d_{\text{emb}}(X \mid \mathcal{G},\mathcal{F}) = d_{\mathcal{F}}(\mathcal{G}(X), X) = \|\mathcal{F}\circ \mathcal{G}(X) - \mathcal{F}(X)\|_2^2.
\end{equation}
Choosing $d_{\text{img}}$ is trickier. In the following subsections, we discuss several choices.

\subsubsection{Hamming Distance}\label{section:l1_distance}
Suppose the image contains $n_p$ pixels (for example, for a grayscale image, this is the product of image width and height). One of the most widely used \citep{shadow_removal_l1_1,shadow_removal_l1_2,pix2pix} distance function for image generation applications is the $L_1$ distance (or, equivalently, the \textit{Hamming distance}) between the ground truth $X$ and generated image $\hat{X}$:
\begin{equation}\label{eq:l1_distance}
    d_H(\hat{X},X) = \frac{1}{n_p}\sum_{i,j,k}|X_{ijk}-\hat{X}_{ijk}|,
\end{equation}
where the sum $\sum_{i,j,k}$ is taken along all pixels on the images (including channels). In contrast to $L_2$ distance, which we define in the following subsection, the Hamming distance encourages less blurring. 

\subsubsection{Euclidean Distance}\label{section:l2_distance}
$L_2$ (Euclidean) distance is also frequently used in image generation applications \citep{shadow_removal_l2,style_transfer}. For grayscale images, it is defined as $\|X-Y\|_{F}$, where $\|\cdot\|_{F}$ denotes the Frobenius norm. For RGB images, similarly to \autoref{eq:l1_distance}, we define the $L_2$ distance as follows: 
\begin{equation}
    d_E(\hat{X},X) = \frac{1}{n_p}\left(\sum_{i,j,k}(X_{ijk}-Y_{ijk})^2\right)^{1/2}.
\end{equation}

\subsubsection{DSSIM}
However, there are multiple ways for a neural network to ``cheat'' in this case. For instance, the neural network might invert background pixels or reduce pixels' intensities since that would not affect embeddings drastically, which in turn will not increase $d_{\text{emb}}$. For this reason, we decided to try using the more advanced method such as a $\mathsf{SSIM}(X,Y)$ (structural similarity index measure) metrics as suggested by \citep{image_loss}. It is defined as:
\begin{equation}
\mathsf{SSIM}(X,Y) = \frac{(2\mu_X\mu_Y + \kappa_1)(2\sigma_{XY} + \kappa_2)}{(\mu_X^2+\mu_Y^2+\kappa_1)(\sigma_X^2+\sigma_Y^2+\kappa_2)},
\end{equation}

where $\mu_X,\mu_Y$ are pixel sample means, $\sigma_X^2,\sigma_Y^2$ are variances, $\sigma_{XY}=\text{cov}[X,Y]$ is a covariance, and $\kappa_1,\kappa_2$ are constants to stabilize the division.

The distance measure, called ``structural dissimilarity'' ($\mathsf{DSSIM}$)\footnote{Note that rigorously speaking, this is not a distance function since triangle inequality is not necessarily satisfied. However, this is not a problem for us if we use this expression as a loss function.}, in turn, is defined as 
\begin{equation}
d_{\text{dssim}}(\hat{X},X) = \frac{1 - \mathsf{SSIM}(\hat{X},X)}{2}.  
\end{equation}

\subsubsection{Sobel Distance}\label{section:sobel_distance}

After experiments, we decided to employ another loss function, which, combined with the $L_1$ loss, performed best on the LFW dataset. Suppose we get an image $X$ as an input. We use two kernels:
\begin{equation}
    K_X = \begin{bmatrix}
        -1 & 0 & 1 \\ -2 & 0 & 2 \\ -1 & 0 & 1
    \end{bmatrix}, \; K_Y = \begin{bmatrix}
        1 & 2 & 1 \\ 0 & 0 & 0 \\ -1 & -2 & -1
    \end{bmatrix}.
\end{equation}
Then, using these two kernels, we find the mask (the sum operation is performed elementwise):
\begin{equation}\label{eq:mask}
    \mathcal{S}(X) = (K_X \star X)^2 + (K_Y \star X)^2.
\end{equation}

Essentially, $\mathcal{S}(X)$ gives a map of regions of $X$ which contain edges.  Finally, we define the distance measure as follows:
\begin{equation}
    d_{\text{sobel}}(\hat{X},X) = d_{H}(\mathcal{S}(X) \odot \hat{X},\mathcal{S}(X) \odot X).
\end{equation}

The difference between this loss and one specified in \autoref{section:l1_distance} is that we account for the loss only in those regions where there are edges since using the pure $L_1$ distance does not restrict the neural network from simply changing the content inside the face without bothering about the shape.

\subsubsection{Combined Distance}\label{section:combined_loss}

Combined loss is just a linear combination of several distances. The best results were achieved by combining the $L_1$ distance (see \autoref{section:l1_distance}) and Sobel distance (see \autoref{section:sobel_distance}):
\begin{equation}
    d_{\text{comb}}(\hat{X},X) = \beta\cdot d_H(\hat{X},X) + (1-\beta)\cdot d_{\text{sobel}}(\hat{X},X),
\end{equation}

where by regulating $\beta$ we can adjust the importance of $d_H$ relative to $d_{\text{sobel}}$. In our experiments, we use $\beta = 0.5$, corresponding to the average between $d_H$ and $d_{\text{sobel}}$.

\subsection{Trainer Network Architecture}

When we finally defined the loss $\ell(X\mid\mathcal{G},\mathcal{F})$, we need to train our generator to minimize this expected loss, that is:
\begin{equation}
\hat{\mathcal{G}} = \argmin_{\mathcal{G}}\mathbb{E}_{X \sim p_{\text{data}}} \ell(X\mid\mathcal{G},\mathcal{F})
\end{equation}

\begin{figure}
    \centering
    \begin{tikzpicture}

    \node[very thick, draw=black, fill=gray, fill opacity=0.2,minimum width=1.75cm, minimum height=4.5cm,dashed,on background layer](targetbox) at (8.5,-1.25) {};
    \node(targetzonelabel)[right=0.6cm of targetbox,rotate=-90,anchor=north]{Output \& Feature vector};
    
    \node[anchor=center,inner sep=0](realimg) at (0,0) {\includegraphics[width=0.1\textwidth]{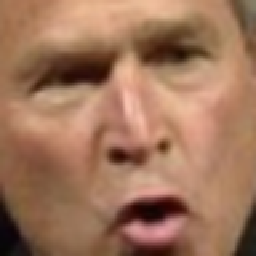}};
    \node[anchor=center,inner sep=0](generatedimg) at (5,0) {\includegraphics[width=0.1\textwidth]{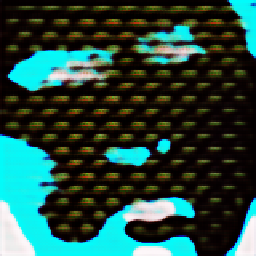}};
    \node[anchor=center,inner sep=0](targetimg) at (8.5,0) {\includegraphics[width=0.1\textwidth]{lfw_6_real.png}};

    \node[draw,shape=rectangle,color=blue!80,fill=blue!2,drop shadow,minimum size=1cm,very thick](gen) at (2.5,0) {\huge $\mathcal{G}$};
    \node[draw,shape=rectangle,color=ForestGreen!80,fill=ForestGreen!2,drop shadow,minimum size=1cm,very thick](emb) at (5,-2.5) {\huge $\mathcal{F}$};

    \node[draw,shape=rectangle,color=black!60,fill=black!2,drop shadow,minimum width=0.5cm, minimum height=1.7cm, very thick](embvecgen) at (6.5,-2.5) {};
    \node[draw,shape=circle,color=black!60,fill=black!10,minimum width=0.3cm, minimum height=0.3cm, very thick](embvecgen1) at (6.5,-2.5) {};
    \node[draw,shape=circle,color=black!60,fill=black!10,minimum width=0.3cm, minimum height=0.3cm, very thick](embvecgen2) at (6.5,-2) {};
    \node[draw,shape=circle,color=black!60,fill=black!10,minimum width=0.3cm, minimum height=0.3cm, very thick](embvecgen3) at (6.5,-3) {};
    \node[draw,shape=rectangle,color=black!60,fill=black!2,drop shadow,minimum width=0.5cm, minimum height=1.7cm, very thick](embvecreal) at (8.5,-2.5) {};
    \node[draw,shape=circle,color=black!60,fill=black!10,minimum width=0.3cm, minimum height=0.3cm, very thick](embvecreal1) at (8.5,-2.5) {};
    \node[draw,shape=circle,color=black!60,fill=black!10,minimum width=0.3cm, minimum height=0.3cm, very thick](embvecreal2) at (8.5,-2) {};
    \node[draw,shape=circle,color=black!60,fill=black!10,minimum width=0.3cm, minimum height=0.3cm, very thick](embvecreal3) at (8.5,-3) {};

    \node(reallabel)[above=0.05 of realimg]{Input $X \in \mathcal{I}$};
    \node(generatedlabel)[above=0.05 of generatedimg]{Generated image $\mathcal{G}(X)$};
    \node[align=center](generatedlabel)[below=0.05 of embvecgen]{Embedding\\$\mathcal{F} \circ \mathcal{G}(X)$};

    \node[align=center,color=blue!80](genmodlabel)[below=0.05 of gen]{Generator\\model};
    \node[align=right,color=ForestGreen!80](embmodlabel)[left=0.15 of emb]{Embedding\\model};

    \path[line width=0.5mm](realimg) edge [-{Stealth[length=3mm]}] node {} (gen);
    \path[line width=0.5mm](gen) edge [-{Stealth[length=3mm]}] node {} (generatedimg);
    \path[line width=0.5mm](generatedimg) edge [-{Stealth[length=3mm]}] node {} (emb);
    \path[line width=0.5mm](emb) edge [-{Stealth[length=3mm]}] node {} (embvecgen);
    \path[line width=0.5mm,above,color=blue!80](generatedimg) edge [{Stealth[length=3mm]}-{Stealth[length=3mm]}] node {$d_{\text{img}}$} (targetimg);
    \path[line width=0.5mm,above,color=ForestGreen!80](embvecgen) edge [{Stealth[length=3mm]}-{Stealth[length=3mm]}] node {$d_{\text{emb}}$} (embvecreal);

    \end{tikzpicture}
    \caption{Trainer Network architecture}
    \label{fig:trainer_network}
\end{figure}

To achieve this, inspired by \citet{inverting}, we create a helper network, which we call a \textit{Trainer Network}. Its architecture is depicted in the \autoref{fig:trainer_network}.

For training, we form the dataset in the following form: the input is an image $X$ while output is a pair of the same image with its embedding $(X,\mathcal{F}(X))$. The trainer network takes an image $X$, generates an image $\mathcal{G}(X)$, and then takes the embedding of this image $\mathcal{F} \circ \mathcal{G}(X)$. It then outputs both values and applies the loss from \autoref{eq:loss_trainer} (since the target value has the same shape). Note that we freeze the embedding network $\mathcal{F}$ and make only $\mathcal{G}$'s weights trainable. 

\section{Implementation}\label{section:implementation}

\subsection{Datasets and Software}

In our research, we used two datasets:
\begin{itemize}
    \item \textbf{MNIST dataset} by \cite{mnist}: dataset, containing $60000$ grayscale images of size $28\times 28$, each with a label from a set $\{0,\dots,9\}$, representing a digit depicted. This dataset is a great starting point for proof-of-concept since testing on it is easy, fast, and insightful.
    \item \textbf{LFW dataset} by \cite{lfw}: dataset consisting of approximately $13000$ RGB face images of size $250 \times 250$ under various poses and lightning conditions. This dataset was used to test that our concept can be successfully transferred to the real biometrics data.
\end{itemize}

\begin{figure*}
\begin{center}
\begin{tabular}{ccccccc}
& \textbf{Image} & \textbf{Label} & \textbf{Image} & \textbf{Label} & \textbf{Image} & \textbf{Label} \\ 
 & \includegraphics[width=0.08\linewidth,valign=m]{mnist_0_real.png} & \Large 0 & \includegraphics[width=0.08\linewidth,valign=m]{mnist_3_real.png} & \Large 3 & \includegraphics[width=0.08\linewidth,valign=m]{mnist_6_real.png} & \Large 6 \\
\textsf{MNIST} & \includegraphics[width=0.08\linewidth,valign=m]{mnist_1_real.png} & \Large 1 & \includegraphics[width=0.08\linewidth,valign=m]{mnist_4_real.png}& \Large 4 & \includegraphics[width=0.08\linewidth,valign=m]{mnist_7_real.png} & \Large 7 \\
& \includegraphics[width=0.08\linewidth,valign=m]{mnist_2_real.png} & \Large 2 & \includegraphics[width=0.08\linewidth,valign=m]{mnist_5_real.png}& \Large 5 & \includegraphics[width=0.08\linewidth,valign=m]{mnist_8_real.png} & \Large 8\\
\\ 
 & \includegraphics[width=0.08\linewidth,valign=m]{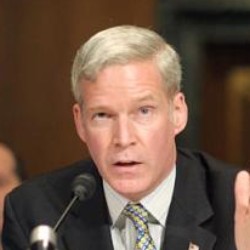} & \scriptsize\textsf{Mark\_Everson} & \includegraphics[width=0.08\linewidth,valign=m]{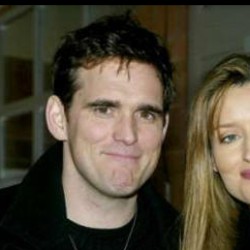}& \scriptsize\textsf{Matt\_Dillon} & \includegraphics[width=0.08\linewidth,valign=m]{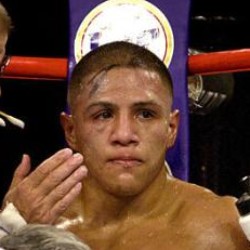} & \scriptsize\textsf{Fernado\_Vargas} \\
\textsf{LFW} & \includegraphics[width=0.08\linewidth,valign=m]{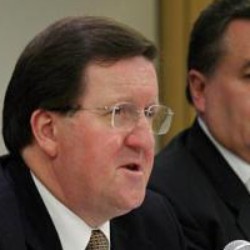} & \scriptsize\textsf{George\_Robertson} & \includegraphics[width=0.08\linewidth,valign=m]{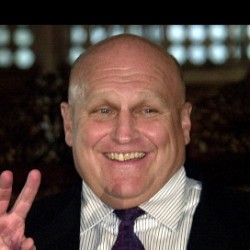}& \scriptsize\textsf{Richard\_Armitage}  & \includegraphics[width=0.08\linewidth,valign=m]{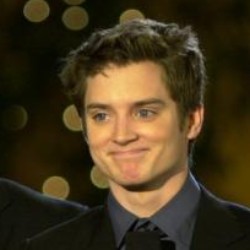} & \scriptsize\textsf{Elijah\_Wood} \\
& \includegraphics[width=0.08\linewidth,valign=m]{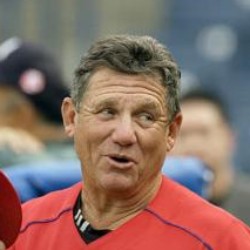} & \scriptsize\textsf{Larry\_Bowa} & \includegraphics[width=0.08\linewidth,valign=m]{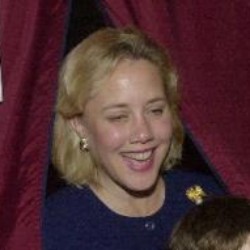}& \scriptsize\textsf{Mary\_Landrieu} & \includegraphics[width=0.08\linewidth,valign=m]{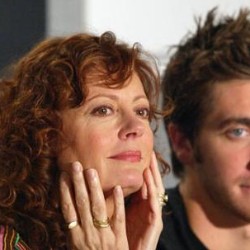} & \scriptsize\textsf{Susan\_Sarandon} 
\end{tabular}
\end{center}
\caption{Example images from \textit{MNIST} \citep{mnist} and \textit{LFW} \citep{lfw} datasets.}
\label{fig:dataset_examples}
\end{figure*}

The example images from both datasets are depicted in \autoref{fig:dataset_examples}.

We used \textit{Python} programming language and \textit{Tensorflow v2.12} (\citet{abadi2016tensorflow}) as the core machine learning platform. We conducted the training and testing on the \textit{MacBook M1}.

\subsection{Embedding Model}\label{section:embedding_implementation}

For the \textit{LFW} dataset, we use the pre-trained \textit{FaceNet} architecture. We decided to employ this architecture since it provides one of the best values of accuracy in the face recognition task: namely, $98.87\%$ for fixed center cropping, and $99.63\%$ for the extra face alignment (see original paper \citep{facenet} for reference). Note that any other embedding neural network might be used, such as \textit{VGGFace} \citep{vggface}, for example.

\begin{figure}
    \centering
    \includegraphics[width=0.8\textwidth]{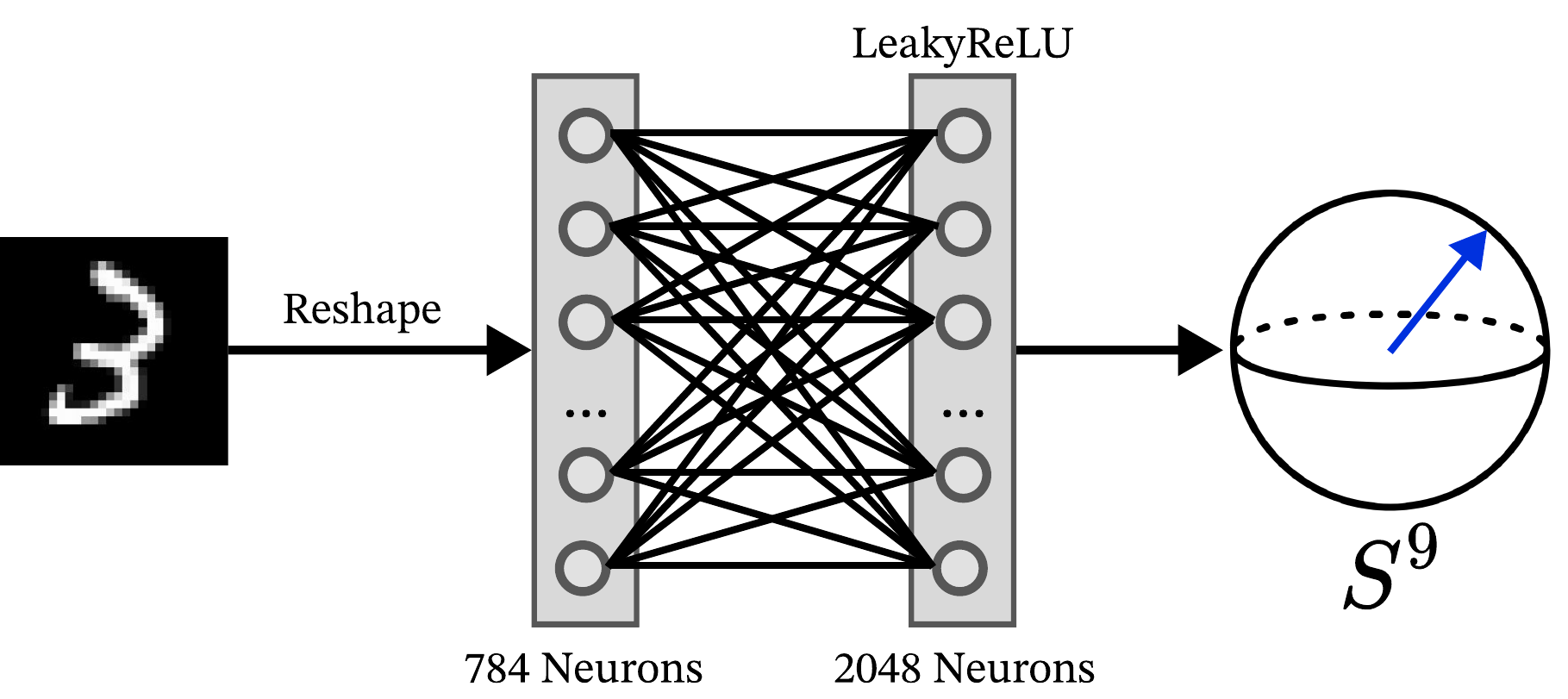}
    \caption{\textit{MNIST} Embedding model architecture. $S^9$ denotes the layer with $10$ neurons which then gets $L_2$ normalized.}
    \label{fig:embedding_mnist}
\end{figure}

For the \textit{MNIST} dataset, we build our own embedding model. We use the architecture specified in \autoref{fig:embedding_mnist}. 

We use the \textsf{LeakyReLU} function defined as $x \mapsto \max\{\alpha x,x\}$ (for $\alpha<1$). We choose $\alpha=0.01$. For the output layer, we do not use an activation function; instead, we normalize the retrieved vector by using $\mathbf{x} \mapsto \frac{\mathbf{x}}{\sqrt{\|\mathbf{x}\|_2^2 + \epsilon}}$ for sufficiently small $0 < \epsilon \ll 1$. As a weight initializer, we use the \textit{He initialization} \citep{he}, which initializes weights according to the normal distribution $\mathcal{N}\left(0,\frac{1}{n_L}\right)$ where $n_L$ is the number of nodes feeding into the layer. We choose our embedding dimensionality to be $m=10$. We then apply the training algorithm described in \autoref{section:triplet_network} using margin $\mu=0.2$ and a learning rate of $\eta = 5 \cdot 10^{-5}$ using \textit{Adam} optimizer \citep{adam}. 

\subsection{Generator Model}\label{section:generator_implementation}

\begin{figure}
    \centering
    \includegraphics[width=0.85\textwidth]{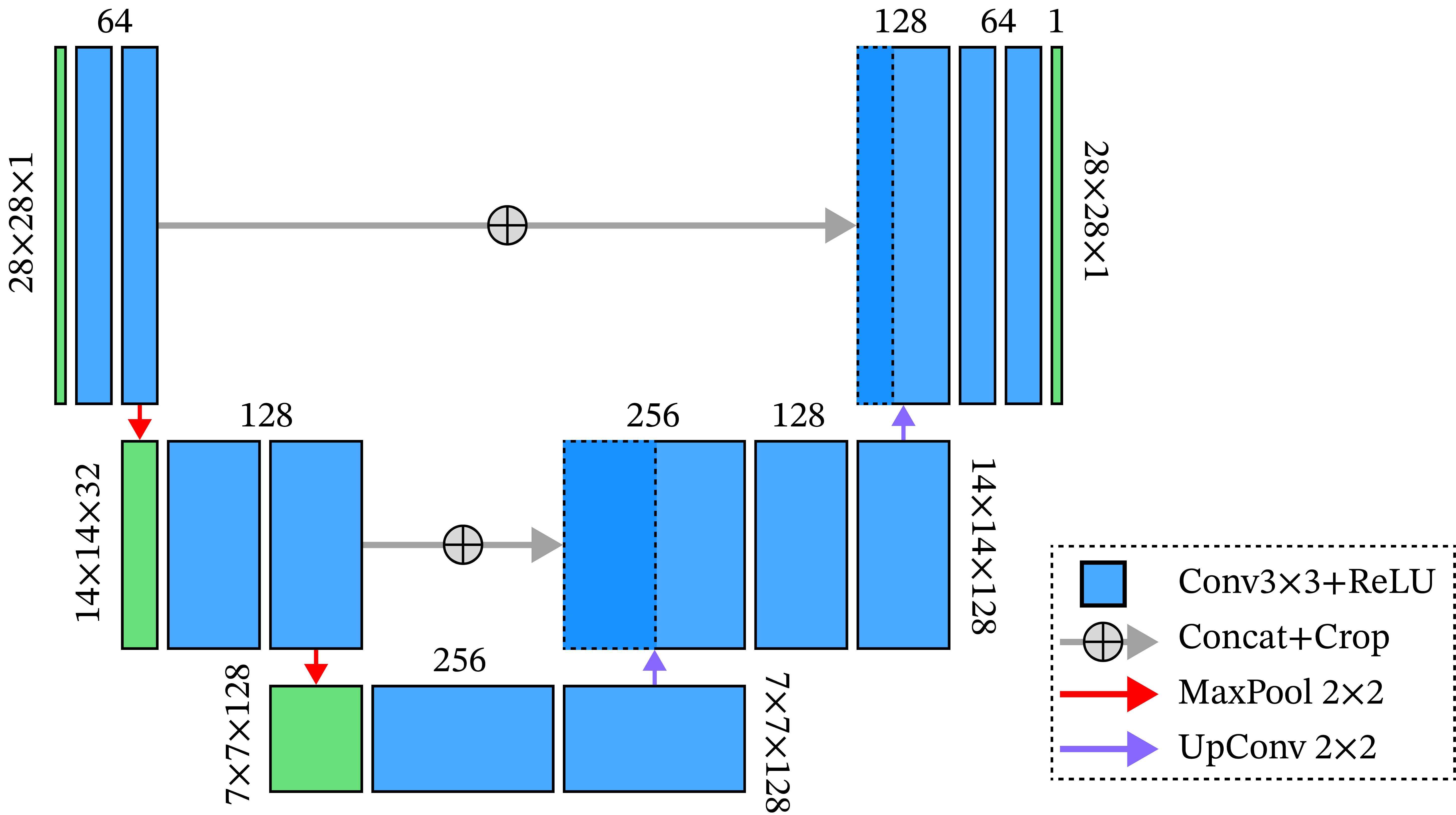}
    \caption{Generator model architecture for the \textit{MNIST} dataset based on \textit{U-Net}.}
    \label{fig:mnist_generator}
\end{figure}

For the generator model, we decided to employ the \textit{U-Net} architecture \citep{unet}, and get the structure specified in \autoref{fig:mnist_generator} for the \textit{MNIST} dataset (architecture for the \textit{LFW} dataset is the same with the only difference in shapes). Similarly to the embedding model from \autoref{section:embedding_implementation}, we use \textit{He} weights initialization, \textsf{LeakyReLU} activation for all convolutional layers except for the last one, and the \textit{sigmoid function} before the output to map pixel values to the interval $(0,1)$. We use batch size of $64$ with a learning rate $\eta=10^{-4}$. Other parameters depend on the dataset:
\begin{itemize}
    \item For the \textit{MNIST} dataset, we use a margin $\alpha=0.3$, $\pi_{\text{emb}}=0.9$, and $L_2$ distance as the loss function (see \autoref{section:l2_distance}).
    \item For the \textit{LFW} dataset, we use a margin $\alpha=0.2,\pi_{\text{emb}}=0.1$, and the combined distance (see \autoref{section:combined_loss}).
\end{itemize}

\section{Results}\label{section:results}

In this section, we analyze the efficacy of the proposed approach after training the neural networks.

\subsection{Image Distance Comparison}

Despite the noticeable changes between the original and generated images, depicted in the \autoref{table:examples}, we still need to provide a quantitative representation of the difference. We will compare images in the following three setups: ``real vs generated same class'', ``real vs real different classes'', and ``generated vs generated different classes''. As a difference metric, we use the $L_2$ distance $d_{E}$ defined in \autoref{section:l2_distance}. We get results specified in \autoref{table:distances}.

\begin{table*}
\begin{center}
\caption{
    \ensuremath{L_2} distances between images of the same digit in three different setups specified as columns. We mark in \textbf{bold} extreme values and highlight in \textcolor{ForestGreen}{\textbf{green}} the best result and in \textcolor{Red}{\textbf{red}} the worst in terms of Real-Gen distances. As can be seen for both \textit{MNIST} and \textit{LFW} datasets, the difference between real and generated images greatly exceeds ``Real-Real'' distances. We use test images (20\% of the whole dataset) from both datasets: approximately 2600 images for \textit{LFW} and 12000 for \textit{MNIST}.
}
\begin{tabular}{c}
\begin{tabular}{cccc} 
 \Xhline{3.5\arrayrulewidth}
 \textbf{Class} & \textbf{Real-Gen} & \textbf{Real-Real} & \textbf{Gen-Gen} \\ [0.5ex] 
 \hline\hline
 {\fontsize{14}{18}\textsf{0}} & $0.791$ & \textbf{0.129} & $0.082$\\ 
 \hline
 {\fontsize{14}{18}\textsf{1}} & \textcolor{ForestGreen}{\textbf{0.886}} & \textbf{0.057} & \textbf{0.031}\\ 
 \hline
 {\fontsize{14}{18}\textsf{2}} & $0.850$ & $0.128$ & $0.108$\\ 
 \hline
 {\fontsize{14}{18}\textsf{3}} & $0.850$ & $0.113$ & \textbf{0.109}\\ 
 \hline
 {\fontsize{14}{18}\textsf{4}} & \textcolor{Red}{\textbf{0.773}} & $0.104$ & $0.049$\\ 
 \hline
 {\fontsize{14}{18}\textsf{5}} & $0.843$ & $0.120$ & $0.106$\\ 
 \hline
 {\fontsize{14}{18}\textsf{6}} & $0.826$ & $0.112$ & $0.058$\\ 
 \hline
 {\fontsize{14}{18}\textsf{7}} & $0.845$ & $0.096$ & $0.062$\\ 
 \hline
 {\fontsize{14}{18}\textsf{8}} & $0.838$ & $0.115$ & $0.087$\\ 
 \hline
 {\fontsize{14}{18}\textsf{9}} & $0.800$ & $0.097$ & $0.048$
 \\
 \Xhline{3.5\arrayrulewidth} \\
\end{tabular} \\
\begin{tabular}{cccc} 
 \Xhline{3.5\arrayrulewidth}
 \textbf{Class} & \textbf{Real-Gen} & \textbf{Real-Real} & \textbf{Gen-Gen} \\ [0.5ex] 
 \hline\hline
 {George W. Bush} & $0.295$ & 0.046 & $0.129$\\ 
 \hline
 {\fontsize{14}{18}Colin Powell} & 0.267 & 0.042 & 0.132\\ 
 \hline
 {\fontsize{14}{18}Tony Blair} & $0.298$ & $0.046$ & $0.122$\\ 
 \hline
 {\fontsize{14}{18}Donald Rumsfeld} & $0.278$ & $0.045$ & 0.120\\ 
 \hline
 {\fontsize{14}{18}Gerhard Schröder} & 0.293 & 0.043 & $0.107$\\ 
 \hline
 {\fontsize{14}{18}Ariel Sharon} & $0.273$ & $0.046$ & \textbf{0.145}\\ 
 \hline
 {\fontsize{14}{18}Hugo Chavez} & \textcolor{Red}{\textbf{0.263}} & $0.041$ & $0.126$\\ 
 \hline
 {\fontsize{14}{18}Junichiro Koizumi} & \textcolor{ForestGreen}{\textbf{0.319}} & $0.045$ & 0.126\\ 
 \hline
 {\fontsize{14}{18}John Ashcroft} & 0.282 & \textbf{0.039} & $0.116$\\ 
 \hline
 {\fontsize{14}{18}Jacques Chirac} & $0.297$ & \textbf{0.051} & \textbf{0.106}
 \\
 \Xhline{3.5\arrayrulewidth}
\end{tabular}
\end{tabular}
\label{table:distances}
\end{center}
\end{table*}

As can be seen, the distance between authentic and generated images have significant values. Consider the \textit{MNIST dataset} as an example: even for pairs of digit \textsf{4} with the minimum value of $0.773$ and especially for digit \textsf{1} with a maximum distance of $0.886$. That highlights that the neural network produced drastically different images in terms of MSE. At the same time, for the \textit{MNIST} dataset, the mean-squared difference remained the same for ``generated vs generated'' pairs, indicating that generation still keeps digits close to each other. In turn, for the \textit{LFW} dataset, the opposite holds: distances between generated faces are significant.

\subsection{Image Encodings Comparison}

To give an intuitive representation of predictions, we apply the PCA \citep{pca} and convert $\mathbb{R}^{m}$ vectors to vectors $\mathbb{R}^3$, which is easy to illustrate on the 3D plot. 

\begin{figure}[htp]
\subfloat[\textit{MNIST} dataset]{
  \includegraphics[trim={100 50 100 50},width=0.48\columnwidth,clip]{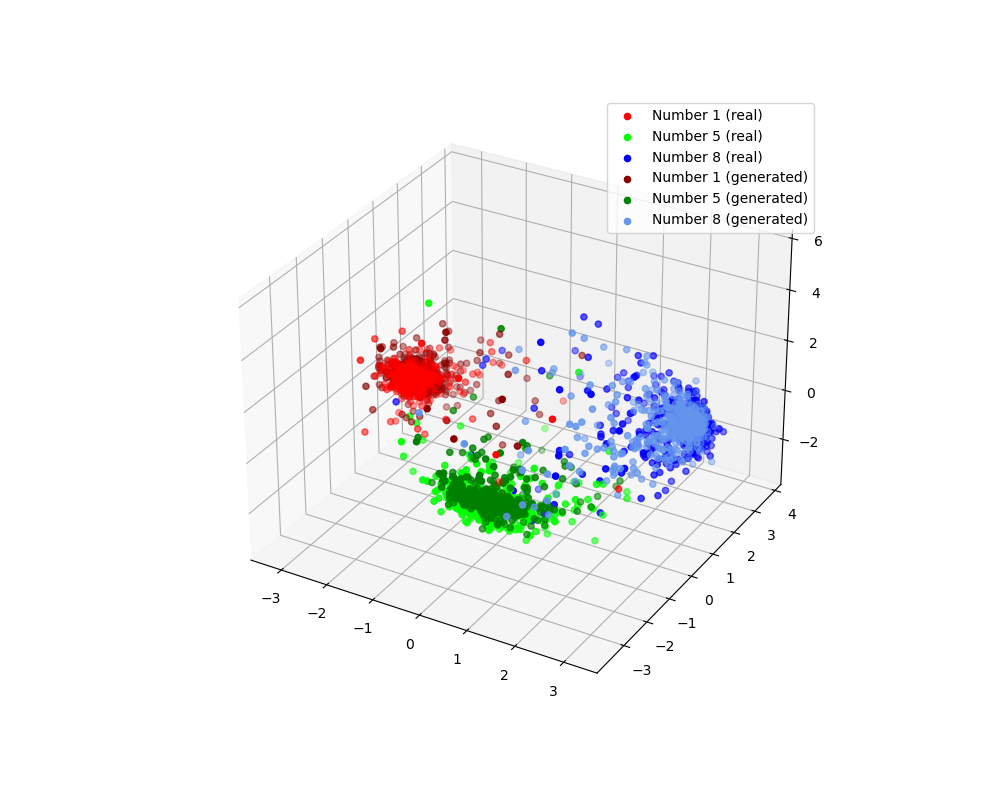}
}
\subfloat[\textit{LFW} dataset]{
  \includegraphics[clip,width=0.48\columnwidth]{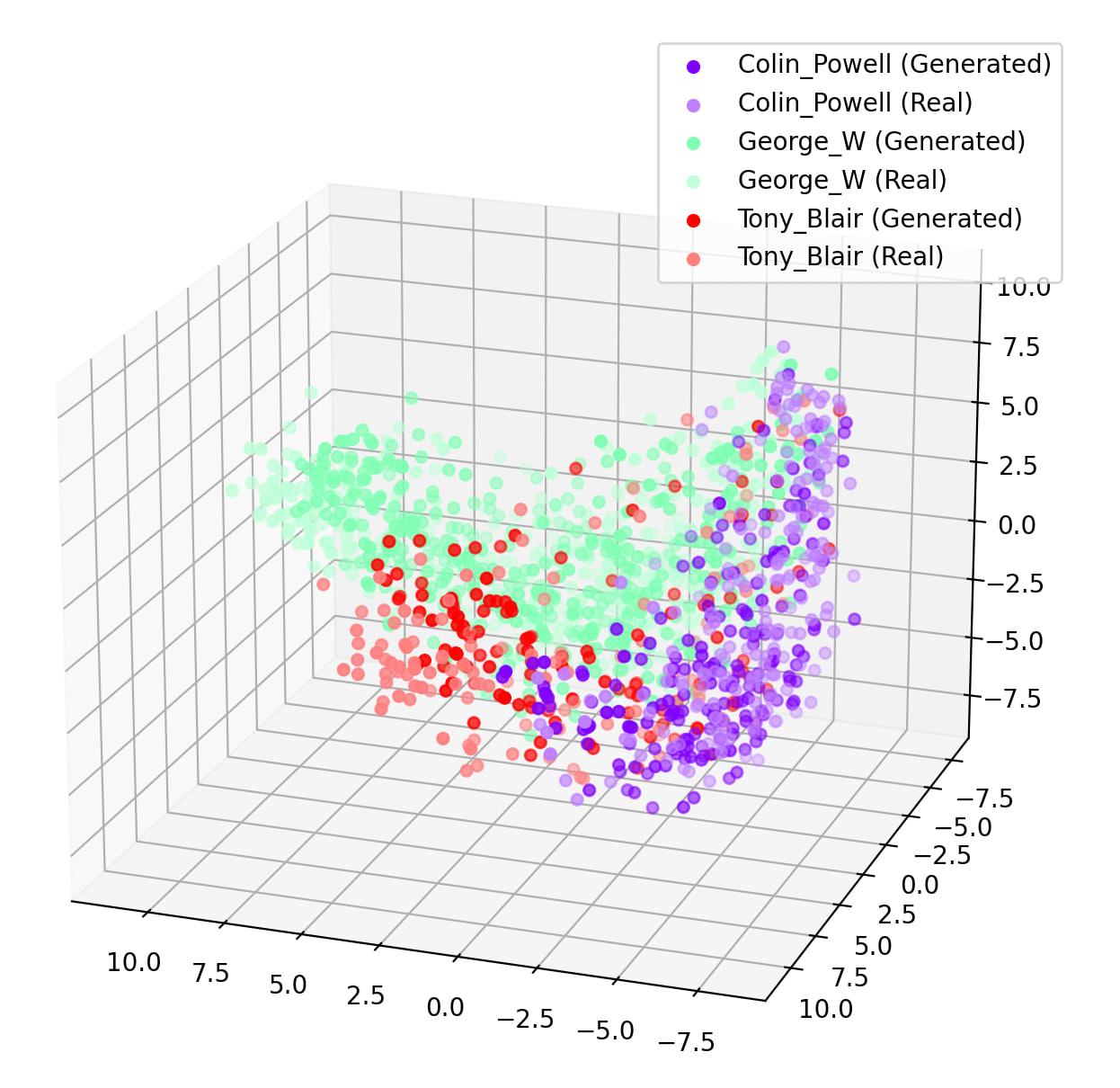}
}
\caption{Embeddings of real and generated images after applying PCA for 3 batches of different classes. We used roughly $300$ embeddings per class for the \textit{MNIST} dataset and roughly $30$ per person for the \textit{LFW} dataset.}
\label{fig:pca_generated}
\end{figure}

That being said, we firstly take a batch of images $\mathsfit{B}:=\{X_i\}_{i=1}^{n_B}$, generate distorted images $\mathsfit{B}_G = \mathcal{G}(\mathsfit{B})$, and then generate two sets of embeddings: $\mathcal{F}(\mathsfit{B})$ and $\mathcal{F}(\mathsfit{B}_G)$. Finally, we apply the PCA to generate three-dimensional representations of $\mathbb{R}^{m}$ embeddings. Results are depicted in the \autoref{fig:pca_generated}. As can be seen, embeddings of the same class almost do not change under the generator transformation and remain close to each other.

\subsection{Dependency on the Margin Parameter}\label{section:margin_tuning}

\begin{figure*}
    \centering
    \includegraphics[width=\textwidth]{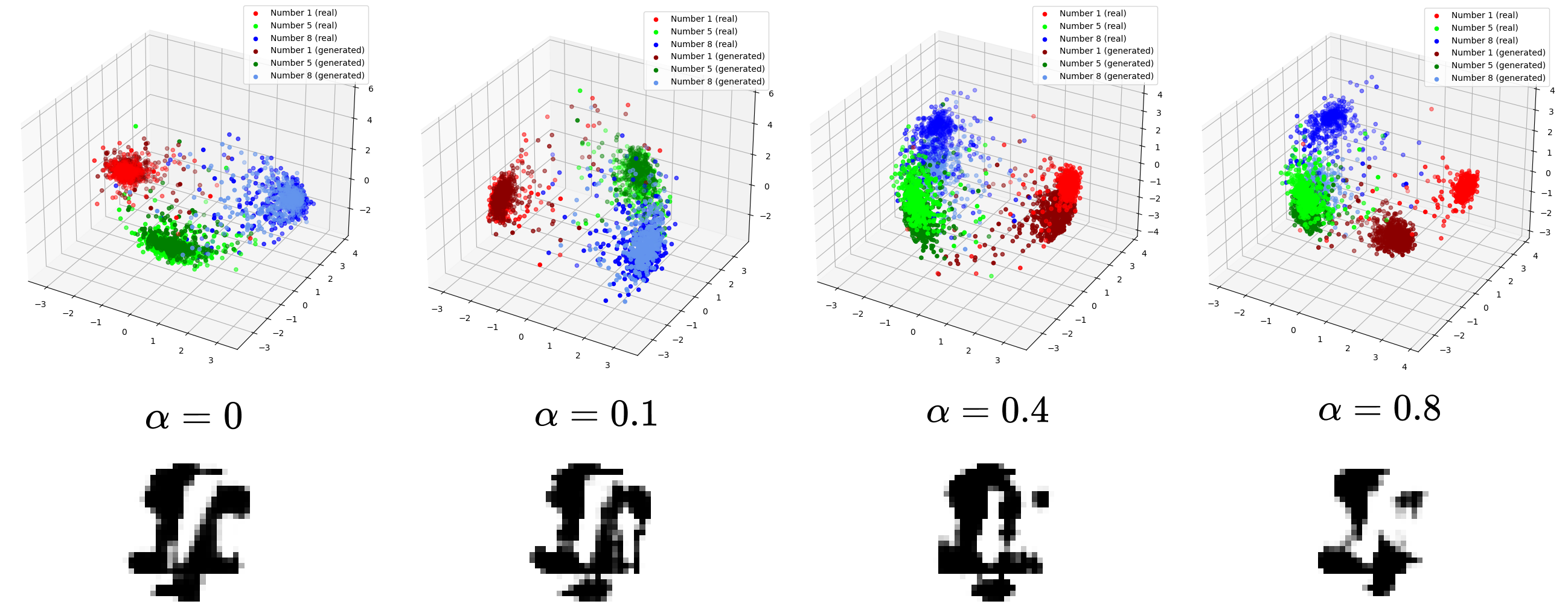}
    \caption{PCA representation of embeddings, corresponding example of a distorted digit of \textsf{1} and margin $\alpha$.}
    \label{fig:margin_tuning}
\end{figure*}

We also tried different values of $\alpha$ to find the best fit. Results for different values of $\alpha$ for the \textit{MNIST} dataset are depicted in the \autoref{fig:margin_tuning}.

As seen, for greater $\alpha$'s, embeddings after generation become more distant from the original ones, but the image distortion is much more considerable. For instance, for $\alpha=0.8$, embeddings of digit \textsf{1} become entirely different from the original ones, and thus this value should not be used for training. $\alpha=0.4$ shows a slight shift of embedding location after generation, but they still remain relatively close. In turn, $\alpha=0$ and $\alpha=0.1$ keep embeddings almost unchanged. In our experiments, empirically, $\alpha=0.2$ provided a sufficient trade-off between embedding and image distances: ``real-gen'' distances remained relatively large (approximately $0.8$ on average). At the same time, the recognition accuracy has not dropped -- see next section for details.

Thus, through extensive empirical evaluations, we found that setting $\alpha=0.2$ provided the best balance between the discriminability of the generated templates (as measured by the EER) and the level of distortion introduced (as assessed by the visual quality of the transformed images). Lower values of $\alpha$ led to insufficient distortion and potential security vulnerabilities, while higher values resulted in excessive distortion that compromised the recognition accuracy. The choice of $\alpha=0.2$ struck an optimal trade-off between these competing objectives, yielding a system with strong security properties and minimal impact on the authentication performance.

\subsection{Mock Recognition System}

In this section, we verify that confusion matrices and ROC curves do not differ significantly if we store distorted images instead of real ones.

For that, we conduct the following experiment: we place distorted images of three classes (for \textit{MNIST} dataset, take \textsf{1},\textsf{2},\textsf{3} for concreteness) in an improvised storage, being simply an in-memory hashmap in our case. Then, we:
\begin{enumerate}
    \item take $1000$ non-distorted images belonging to these three classes and try entering into the ``system'';
    \item take $1000$ non-distorted images of any other three classes (except for previously chosen triplet) and try logging in.
\end{enumerate}

We expect the former to be a successful login attempt while the latter to be an invalid authorization. We then build confusion matrices by providing a number of \textsf{TP}s (true positives), \textsf{TN}s (true negatives), \textsf{FP}s (false positives), and \textsf{FN}s (false negatives). We then calculate the following metrics:
\begin{gather}
\mathsf{Precision} = \frac{\mathsf{TP}}{\mathsf{TP}+\mathsf{FP}}, \; \mathsf{Recall} = \frac{\mathsf{TP}}{\mathsf{TP}+\mathsf{FN}}, \\ F_1 = \frac{2 \times \mathsf{Precision} \times \mathsf{Recall}}{\mathsf{Precision} + \mathsf{Recall}}.
\end{gather}

\begin{figure}[htp]
\centering
\subfloat[\textit{MNIST} dataset]{
  \includegraphics[width=0.45\columnwidth,clip]{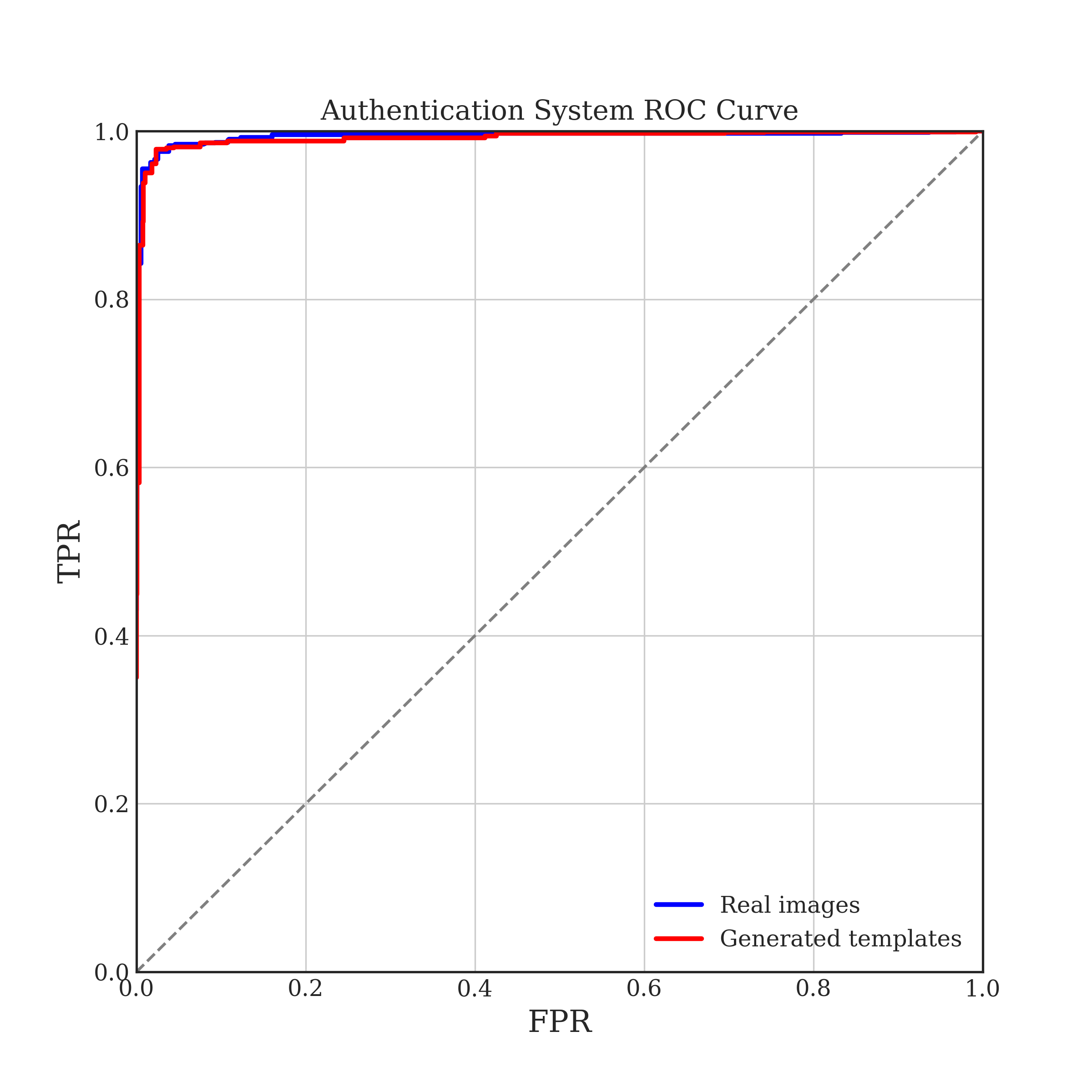}
}
\subfloat[\textit{LFW} dataset]{
  \includegraphics[clip,width=0.45\columnwidth]{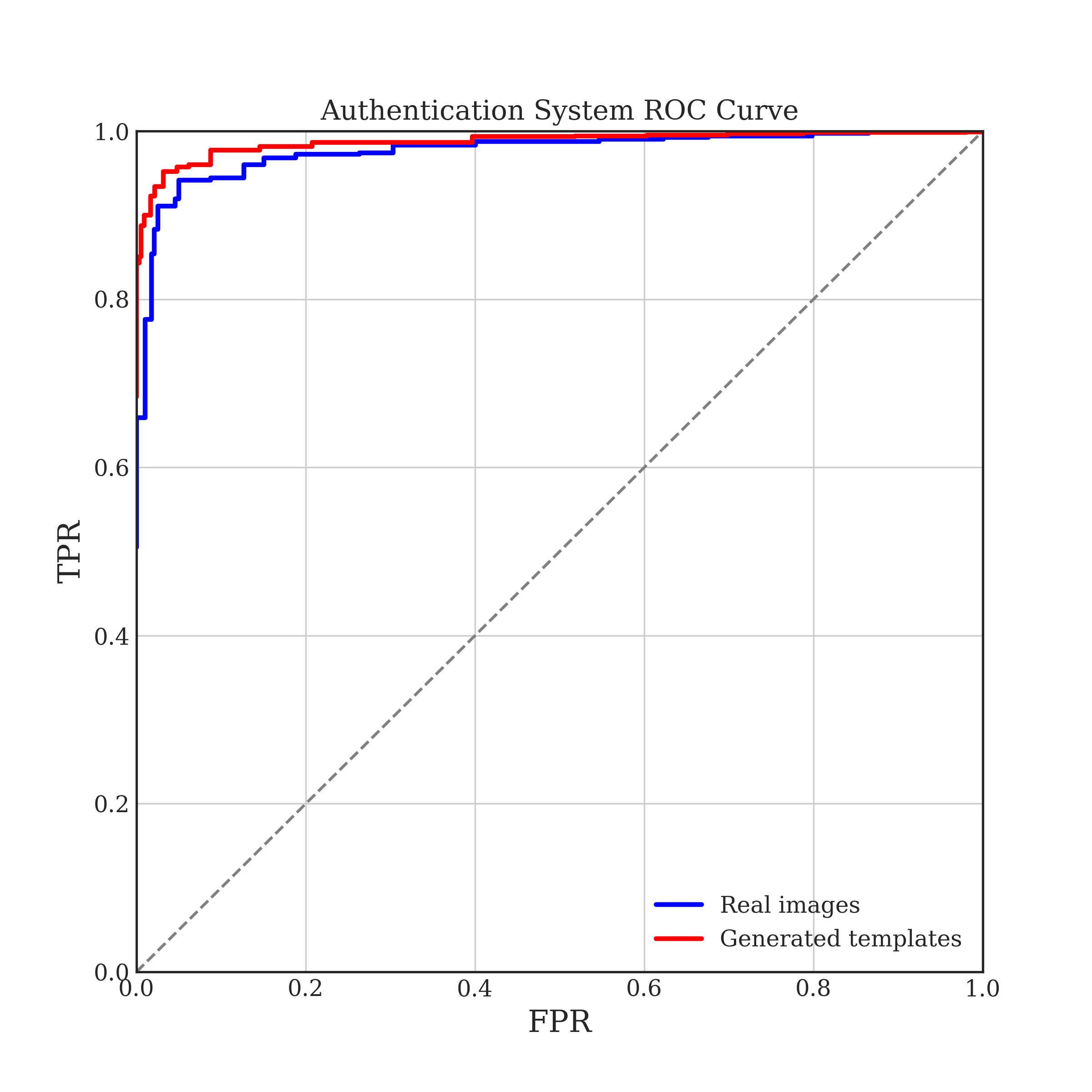}
}
\caption{ROC curve for a mock authentication system using \textbf{(a)} \textit{MNIST} and \textbf{(b)} \textit{LFW} datasets. \textcolor{Red}{Red} color represents the curve for a case where we store distorted images in the storage while \textcolor{blue}{blue} color corresponds to storing real images.}
\label{fig:roc}
\end{figure}

We take $1000$ different values for a threshold $\tau$ in range $[0,4]$ and classify images $X,Y$ to be of the same class if $d_{\mathcal{F}}(X,Y) < \tau$ and of different ones otherwise. We then chose a threshold providing the best $F_1$ score and built the corresponding confusion matrix. We use 80\% of images for training and 20\% for verifying the results, corresponding to approximately 12000 images in the MNIST dataset and almost 2600 photos in the LFW dataset. We get results depicted in the \autoref{table:confusion_matrices} and ROC curves shown in \autoref{fig:roc}. 

\begin{table*}
\begin{center}
\caption{
    Confusion matrices and metric values for authentication system with(a) and without(b) distorting original inputs.
}
\begin{tabular}{ cc }
\multicolumn{2}{c}{\textbf{MNIST Dataset}} \\

(a) Without distortion & (b) With distortion \\  

\begin{tabular}{ cc|c|c } 
  \multicolumn{4}{c}{\textbf{Prediction}} \\ 
\multirow{4}{*}{\rotatebox{90}{\textbf{Actual}}} & & Positive & Negative \\ \cline{2-4}
    & Positive & \cellcolor{blue!25}{\textbf{1174}} & 26 \\ \cline{2-4}
    & Negative & 29 & \cellcolor{blue!25}\textbf{{1171}}  \\ \cline{2-4} \\
    & Precision & \multicolumn{2}{c}{97.59\%}
    \\ \cline{2-4} 
    & Recall & \multicolumn{2}{c}{97.83\%}
    \\ \cline{2-4} 
    & $F_1$ score & \multicolumn{2}{c}{97.71\%} 
\end{tabular} &  

\begin{tabular}{ cc|c|c } 
  \multicolumn{4}{c}{\textbf{Prediction}} \\ 
\multirow{4}{*}{\rotatebox{90}{\textbf{Actual}}} & & Positive & Negative \\ \cline{2-4}
    & Positive & \cellcolor{blue!25}{\textbf{1171}} & 29 \\ \cline{2-4}
    & Negative & 32 & \cellcolor{blue!25}\textbf{{1168}}  \\ \cline{2-4} \\
    & Precision & \multicolumn{2}{c}{97.34\% (\textcolor{Red}{$\downarrow 0.25\%$})}
    \\ \cline{2-4} 
    & Recall & \multicolumn{2}{c}{97.58\% (\textcolor{Red}{$\downarrow 0.25\%$})}
    \\ \cline{2-4} 
    & $F_1$ score & \multicolumn{2}{c}{97.46\% (\textcolor{Red}{$\downarrow 0.25\%$})}
\end{tabular} 
\\ \\
\multicolumn{2}{c}{\textbf{LFW Dataset}}
\\
\begin{tabular}{ cc|c|c } 
  \multicolumn{4}{c}{\textbf{Prediction}} \\
\multirow{4}{*}{\rotatebox{90}{\textbf{Actual}}} & & Positive & Negative \\ \cline{2-4}
    & Positive & \cellcolor{blue!25}{\textbf{1130}} & 70 \\ \cline{2-4}
    & Negative & 80 & \cellcolor{blue!25}\textbf{{1120}}  \\ \cline{2-4} \\
    & Precision & \multicolumn{2}{c}{93.34\%}
    \\ \cline{2-4} 
    & Recall & \multicolumn{2}{c}{94.17\%}
    \\ \cline{2-4} 
    & $F_1$ score & \multicolumn{2}{c}{93.78\%} 
\end{tabular} &  

\begin{tabular}{ cc|c|c } 
  \multicolumn{4}{c}{\textbf{Prediction}} \\ 
\multirow{4}{*}{\rotatebox{90}{\textbf{Actual}}} & & Positive & Negative \\ \cline{2-4}
    & Positive & \cellcolor{blue!25}{\textbf{1142}} & 58 \\ \cline{2-4}
    & Negative & 57 & \cellcolor{blue!25}\textbf{{1143}}  \\ \cline{2-4} \\
    & Precision & \multicolumn{2}{c}{95.25\% (\textcolor{ForestGreen}{$\uparrow 1.91\%$})}
    \\ \cline{2-4} 
    & Recall & \multicolumn{2}{c}{95.17\% (\textcolor{ForestGreen}{$\uparrow 1.00\%$})}
    \\ \cline{2-4} 
    & $F_1$ score & \multicolumn{2}{c}{95.21\% (\textcolor{ForestGreen}{$\uparrow 1.43\%$})}
\end{tabular}

\end{tabular}
\vspace{10px}
\label{table:confusion_matrices}
\end{center}
\end{table*}

As we can see, accuracy metrics do not differ significantly under the image distortion and therefore we have successfully achieved our goal. Moreover, the distortion-generated technique even slightly outscored the non-distortive approach. 

\subsection{Limitations}

Certainly, during the training process, we encountered numerous issues and obstacles, some of which are depicted in the \autoref{table:challenges} together with the causes. Some of them include:
\begin{itemize}
    \item \textbf{Vanishing or exploding gradients}: the generator model produces the same blank image regardless of the input. 
    \item \textbf{Highlighting the contours without concealing effect}: the generator model ``cheats'' by not changing the contours but instead changing the content inside them. This results in an image, from which it is easy to recognize the face.
    \item \textbf{Changing the color gamma}: the neural network simply changes the image's gamma, which surely does not conceal the face.
\end{itemize}

\begin{table*}
\begin{center}
\caption{Three primary challenges when training the generator model: vanishing gradients, highlighting the contours, and changing the color gamma, and corresponding examples with possible causes.}
\begin{tabular}{cccccc}
\Xhline{3.5\arrayrulewidth}
\multirow{2}{*}{\textbf{Problem}} & \multicolumn{2}{c}{\textbf{MNIST}} & \multicolumn{2}{c}{\textbf{LFW}} & \multirow{2}{*}{\textbf{Possible Cause}} \\ 
& Real & Generated & Real & Generated & \\
\makecell{Vanishing\\or exploding \\ gradients} & \includegraphics[width=.12\linewidth,valign=m]{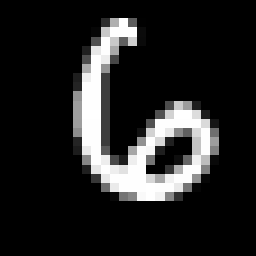} & \includegraphics[width=.12\linewidth,valign=m]{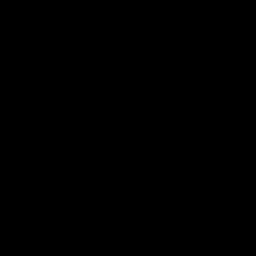} & \includegraphics[width=.12\linewidth,valign=m]{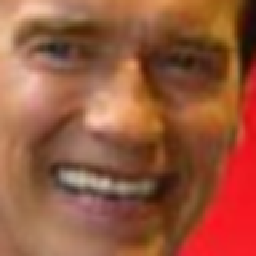} & \includegraphics[width=.12\linewidth,valign=m]{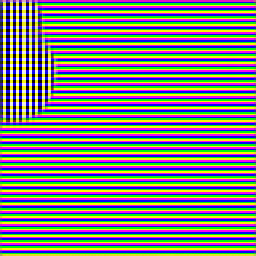} & \makecell[l]{1. Too large learning rate. \\ 2. Too small $\pi_{\text{emb}}$ or \\ too large $\alpha$: ignoring \\ preserving embeddings.} \\\\
\makecell{Highlighting the\\contours without\\concealing effect} & \includegraphics[width=.12\linewidth,valign=m]{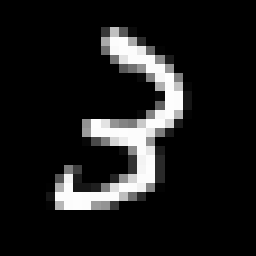} & \includegraphics[width=.12\linewidth,valign=m]{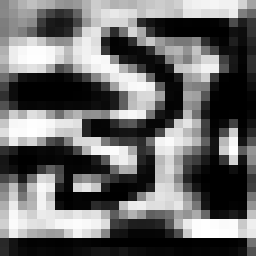} & \includegraphics[width=.12\linewidth,valign=m]{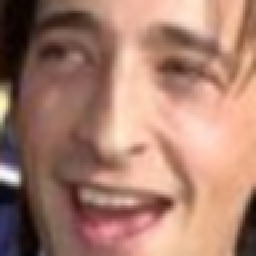} & \includegraphics[width=.12\linewidth,valign=m]{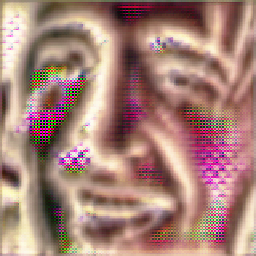} & \makecell[l]{1. Too large $\pi_{\text{emb}}$: \\ focusing too much \\ on saving embeddings.\\
2. Too small $\alpha$.\\
3. Typically happens \\ for SSIM loss.} \\
\\
\makecell{Changing the\\color gamma} & \includegraphics[width=.12\linewidth,valign=m]{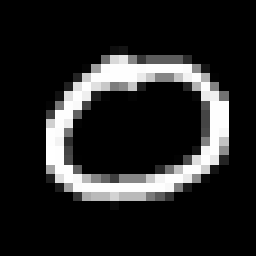} & \includegraphics[width=.12\linewidth,valign=m]{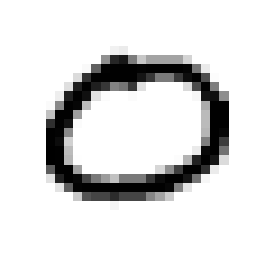} & \includegraphics[width=.12\linewidth,valign=m]{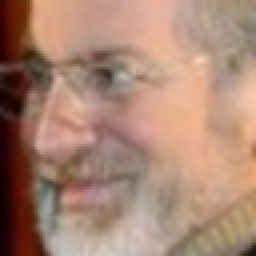} & \includegraphics[width=.12\linewidth,valign=m]{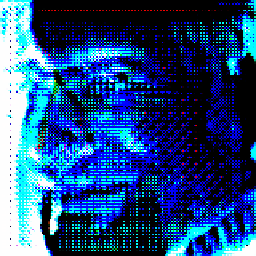} & \makecell[l]{1. Bad balance \\ between $\pi_{\text{emb}}$,\\ learning rate, and $\alpha$. \\ 2. Typically happens \\for $L_1$ or $L_2$ loss.} \\
\Xhline{3.5\arrayrulewidth}
\end{tabular}
\label{table:challenges}
\end{center}
\end{table*}

\section{Comparison to other research}\label{section:comparison}

In this section, we delve deeper into the comparative analysis of our Non-Distortive Cancelable Biometrics system with existing notable works in the field of biometric security. The focus is on understanding how our approach aligns with or diverges from these established methods, particularly in terms of performance metrics like the Equal Error Rate (EER).

In the comparative analysis presented in \autoref{tab:comparison}, we juxtapose the EER of various biometric authentication systems, including our own, against a backdrop of diverse datasets and biometric modalities. This table serves as a crucial benchmark, allowing us to contextualize our Non-Distortive Cancelable Biometrics system within the broader landscape of biometric security research.

\begin{table*}[ht]
\centering
\caption{Comparative Analysis of Biometric Authentication Systems}
\label{tab:comparison}
\begin{tabular}{l l c}
\Xhline{3.5\arrayrulewidth}
\textbf{Source} & \textbf{Type of Images, Dataset} & \textbf{EER, (\%)} \\
\hline
 \citet{Yang_Wang_Kang_Johnstone_Bedari_2022} & Fingerprint, \textit{FVC2002} & 0.5 -- 4.5 \\
 \citet{Yang_Wang_Kang_Johnstone_Bedari_2022} & Fingerprint, \textit{FVC2004} & 2.7 -- 6.3 \\
 \citet{Yang_Wang_Kang_Johnstone_Bedari_2022} & Face, \textit{LFW} & 1.9 \\
 \citet{Yang_Wang_Kang_Johnstone_Bedari_2022} & Fingerprint, \textit{FVC2002} & 7.6 -- 9.4 \\
\citet{Yang_Wang_Kang_Johnstone_Bedari_2022} & Fingerprint, \textit{FVC2004} & 15.6 \\
 \citet{Kaur_Khanna_2020} & Face, \textit{CASIA} & 2.2 -- 9.3 \\
 \citet{Yang_Wang_Shahzad_Zhou_2021} & Fingerprint, \textit{FVC2002} & 1.0 -- 4.0 \\
 \citet{Yang_Wang_Shahzad_Zhou_2021} & Fingerprint, \textit{FVC2004} & 11 \\
 \citet{Wang_Deng_Hu_2017} & Fingerprint, \textit{FVC2002} & 1.0 -- 5.2 \\
 \citet{Wang_Deng_Hu_2017} & Fingerprint, \textit{FVC2004} & 13.3 \\
\textbf{Our Work} & Numbers, \textit{MNIST} & 2.5 \\
\textbf{Our Work} & Face, \textit{LFW} & 4.8 \\
\Xhline{3.5\arrayrulewidth}
\end{tabular}
\end{table*}

The work of \citet{Lee_Teoh_Uhl_Liang_Jin_2021} in multimodal biometric systems stands out for its impressive EER range, particularly in fingerprint recognition on the \textit{FVC2002} and \textit{FVC2004} datasets, and facial recognition on the \textit{LFW} dataset. Their EERs, spanning from as low as 0.5\% to 6.3\%, underscore the efficacy of leveraging multiple biometric modalities. This multimodal approach, by integrating diverse biometric data, enhances the overall system robustness, a feature that our system aims to emulate in a single-modality context.

\citet{Yang_Wang_Kang_Johnstone_Bedari_2022} present a higher EER for fingerprint recognition, particularly on the \textit{FVC2004} dataset, where the EER peaks at 15.6\%. This elevated rate could be indicative of the challenges inherent in the dataset or perhaps limitations in the methodological approach they employed. In contrast, our system, while not directly comparable due to different modalities, shows a more favorable EER of 4.8\% for facial recognition on the LFW dataset, suggesting a more robust performance in handling biometric variability.

\citet{Kaur_Khanna_2020} explore facial biometrics using the \textit{CASIA} dataset, with their EER ranging from 2.2\% to 9.3\%. The broad range of their EER might reflect the varying complexities within the dataset and the adaptability of their system to different facial features. Our system, while tested on a different facial dataset (\textit{LFW}), demonstrates a competitive edge with a consistent EER, highlighting its potential for reliable performance across diverse facial data.

The studies by \citet{Yang_Wang_Shahzad_Zhou_2021} and \citet{Wang_Deng_Hu_2017} focus on fingerprint biometrics, with EERs that offer a balanced perspective on security and usability. \citet{Yang_Wang_Shahzad_Zhou_2021} report EERs ranging from 1.0\% to 4.0\% for \textit{FVC2002} and 11\% for \textit{FVC2004}, while \citet{Wang_Deng_Hu_2017} present EERs from 1.0\% to 5.2\% for \textit{FVC2002} and 13.3\% for \textit{FVC2004}. These results, though specific to fingerprint biometrics, provide valuable insights into the efficacy of different biometric processing techniques, which are instrumental in guiding our approach to facial biometric authentication.

Our work, with an EER of 2.5\% on the \textit{MNIST} dataset and 4.8\% on the \textit{LFW} dataset, demonstrates a promising balance between security and usability. The \textit{MNIST} dataset, though less complex, serves as a foundational testbed, validating the core principles of our approach. The \textit{LFW} dataset, more representative of real-world scenarios, further affirms the robustness and applicability of our system in a practical context.

In summary, our comparative analysis not only situates our Non-Distortive Cancelable Biometrics system within the current state of biometric security research but also highlights its potential as a competitive and innovative solution. 

\section{Discussions}\label{section:Discussions}

In this section, we delve deeper into the comparative analysis of our Non-Distortive Cancelable Biometrics system with existing notable works in the field of biometric security. The focus is on understanding how our approach aligns with or diverges from these established methods, particularly in terms of performance metrics like the EER.

The comparative analysis presented in \autoref{tab:comparison} juxtaposes the EER of various biometric authentication systems, including our own, against a backdrop of diverse datasets and biometric modalities. This table serves as a crucial benchmark, allowing us to contextualize our Non-Distortive Cancelable Biometrics system within the broader landscape of biometric security research.

Our work, with an EER of 4.8\% on the \textit{LFW} facial dataset, demonstrates a promising balance between security and usability. The \textit{LFW} dataset, being representative of real-world scenarios with unconstrained facial images, affirms the robustness and practical applicability of our system. This result aligns favorably with the state-of-the-art cancelable biometric systems for face recognition, such as the work by \citet{Yang_Wang_Kang_Johnstone_Bedari_2022}, which reports an EER of 1.9\% on the same \textit{LFW} dataset. The slight variation in performance can be attributed to differences in feature extraction and transformation techniques, as well as the inherent trade-off between privacy and accuracy in our non-distortive approach.

It's noteworthy that our system's performance remains consistent across different biometric modalities. The EER of 2.5\% on the \textit{MNIST} handwritten digit dataset further validates the generalizability of our approach. While the \textit{MNIST} dataset serves as a preliminary testbed, the low error rate underscores the robustness of our feature preservation mechanism and the efficacy of the proposed distortion method.

Broadening the comparative scope, we observe that our results are highly competitive with cancelable biometric systems designed for other modalities, such as fingerprints. The works by \citet{Wang_Deng_Hu_2017}, \citet{Yang_Wang_Kang_Johnstone_Bedari_2022}, and  \citet{Yang_Wang_Shahzad_Zhou_2021} report EERs ranging from 0.5\% to 15.6\% on the \textit{FVC2002} and \textit{FVC2004} fingerprint datasets. The fact that our system's performance falls within this range, despite the inherent differences in biometric characteristics and dataset complexities, underscores the potential of our non-distortive paradigm.

However, it's crucial to acknowledge that direct comparisons across different biometric modalities and datasets are not always straightforward. Factors such as sensor quality, population demographics, and environmental conditions can significantly influence the performance metrics. Moreover, the specific security requirements and privacy regulations associated with each application domain may dictate different trade-offs between the degree of distortion and the recognition accuracy.

Despite these challenges, the comparative analysis in \autoref{tab:comparison} provides a valuable perspective on the positioning of our work within the biometric security landscape. The competitive EERs across facial and handwritten digit datasets, coupled with the novel non-distortive cancelable biometrics paradigm, underscore the potential of our approach to reshape the practice of biometric authentication.

Naturally, a key avenue for future work is to extend the experimental validation to a broader spectrum of biometric characteristics, including fingerprints. While the current study has laid the theoretical and empirical foundations, larger-scale trials across demographics, environmental conditions, and attack scenarios are necessary to reinforce the real-world applicability. Longitudinal studies can also shed light on template ageing effects and the need for re-enrollment protocols.

On the algorithmic front, there is ample room for refining the feature extraction, embedding, and distortion generation components. Adversarial learning techniques, potentially allowing joint optimization of the embedding and distortion networks, are a particularly promising direction. Advances in explainable AI could also enable greater interpretability of the learned representations and failure modes.

\section{Conclusion}\label{section:conclusion}

This paper has presented a novel approach to biometric security that maintains the integrity of original biometric data while ensuring robust security and privacy. The experimental results, leveraging the \textit{MNIST} and \textit{LFW} datasets and advanced deep learning algorithms, have demonstrated the feasibility and effectiveness of this innovative system.

Key findings include:
\begin{itemize}
    \item \textbf{Feasibility of Non-Distortive Approach.} The experiments have successfully shown that it is possible to generate cancelable biometric templates that retain high similarity in AI metrics while appearing significantly different in traditional metrics. This finding is crucial as it validates the core premise of the Non-Distortive Cancelable Biometrics system.
    \item \textbf{AI-Driven Metric Similarity.} AI algorithms, particularly convolutional neural networks, have proven effective in maintaining metric similarity between the original and transformed biometric data. This aspect underscores the potential of AI in enhancing biometric security.
    \item \textbf{Security and Privacy.} The system's ability to generate non-invertible and non-replicable biometric templates addresses significant concerns regarding data security and user privacy in traditional biometric systems.
    \item \textbf{Operational Flexibility.} The adaptability of the system to various biometric modalities and its scalability across different platforms.
\end{itemize}
The promising results of this study pave the way for further research and development in this field. Future work could focus on:
\begin{itemize}
    \item \textbf{Enhancing AI Algorithms.} Continuous improvement of the AI algorithms for more nuanced feature extraction and comparison.
    \item \textbf{Expanding Biometric Modalities.} Exploring the application of this system to other biometric data types such as voice recognition or gait analysis.
    \item \textbf{Real-World Implementation.} Testing the system in real-world scenarios to assess its practicality and performance under varied conditions.
\end{itemize}
In conclusion, the Non-Distortive Cancelable Biometrics system represents a significant step forward in biometric security. Its ability to balance security, privacy, and operational efficiency sets a new benchmark for future biometric systems. The insights from this research contribute substantially to the ongoing discourse in biometric technology, offering a viable and innovative solution to the challenges faced in this rapidly evolving field.

\section*{CRediT authorship contribution statement}

Dmytro Zakharov: Methodology, Writing–original draft. Oleksandr Kuznetsov: Conceptualization \& Data curation,  Writing – review \& editing. Emanuele Frontoni:  Investigation \& Supervision. 

\section*{Declaration of competing interest} 

The authors declare that they have no known competing financial interests or personal relationships that could have appeared to influence the work reported in this paper. 

\section*{Data availability} 

Data will be made available on request.

\section*{Acknowledgement}

\begin{itemize}
    \item This project has received funding from the European Union’s Horizon 2020 research and innovation programme under the Marie Skłodowska-Curie grant agreement No. 101007820 - TRUST. This publication reflects only the author’s view and the REA is not responsible for any use that may be made of the information it contains.
    \item This research was funded by the European Union – NextGenerationEU under the Italian Ministry of University and Research (MIUR), National Innovation Ecosystem grant ECS00000041-VITALITY-CUP D83C22000710005.

\end{itemize}
 
\bibliographystyle{elsarticle-harv} 
\bibliography{cas-refs}

\end{document}